%% file: colm2026_conference.tex
\documentclass{article} 
\usepackage[final]{colm2026_conference}

\usepackage{microtype}
\usepackage{hyperref}
\usepackage{url}
\usepackage{booktabs}
\usepackage{pgfplots}
\usepackage{latexsym}
\usepackage{enumitem}
\usepackage{multirow}
\usepackage{caption}
\usepackage{placeins}
\usepackage{makecell}
\usepackage{amsfonts}
\usepackage{xspace}
\usepackage{graphicx}
\usepackage{xcolor}
\usepackage{array}
\usepackage{dashbox} 
\usepackage{microtype}
\usepackage{amsmath}
\usepackage{wrapfig}
\usepackage{subcaption}

\usepackage{wrapfig}
\usepackage[table]{xcolor}
\usepackage{colortbl}
\usepackage{arydshln}

\usepackage{lineno}

\definecolor{darkblue}{rgb}{0, 0, 0.5}
\hypersetup{colorlinks=true, citecolor=darkblue, linkcolor=darkblue, urlcolor=darkblue}

\usepackage{titlesec}
\usepackage[font=small,skip=3pt]{caption}

\titlespacing*{\section}
  {0pt}{0.7ex plus 0.3ex minus 0.1ex}{0.3ex plus 0.1ex}

\titlespacing*{\subsection}
  {0pt}{0.3ex plus 0.1ex minus 0.1ex}{0.2ex plus 0.1ex}

\titlespacing*{\subsubsection}
  {0pt}{0.2ex plus 0.1ex minus 0.1ex}{0.1ex plus 0.1ex}

\titlespacing*{\paragraph}
  {0pt}{0.1ex plus 0.0ex minus 0.0ex}{0.5em}

\input{commands}

\title{Clarify or Answer: Reinforcement Learning for Agentic VQA with Context Under-specification}

\author{
\textbf{Zongwan Cao}\textsuperscript{1}\thanks{Equal contribution}\quad
\textbf{Bingbing Wen}\textsuperscript{1}\footnotemark[1]\quad
\textbf{Lucy Lu Wang}\textsuperscript{1,2} \\
\textsuperscript{1}University of Washington \\
\textsuperscript{2}Allen Institute for AI \\
\texttt{\{zongwanc, bingbw, lucylw\}@uw.edu}
}

\begin{document}

\ifcolmsubmission
\linenumbers
\fi

\maketitle

\vspace{-1mm}
\begin{abstract}
\vspace{-1mm}
Real-world visual question answering (VQA) may be \emph{context-dependent}: an image-question pair could be under-specified, such that the correct answer depends on external information that is not observable in the image. In such cases, directly answering can lead to confident but incorrect predictions. We propose Clarify-or-Answer (CoA), an ask-or-answer agent that separately models the decision to ask or answer, and what to ask if needed. CoA first determines whether clarification is necessary; if so, it generates one or more clarification questions to iteratively enrich the context until ambiguity is resolved. To validate this framework, we introduce \dataset, a dataset of ambiguous VQA questions along with a contrast set of non-ambiguous questions. We further introduce \method (Clarification Reasoning), a reinforcement learning approach that optimizes clarification question generation with multiple reward signals encouraging well-formed, focused, non-trivial questions that resolve ambiguity. Across three VLMs and three datasets, CoA achieves consistent improvements at both the module and system levels, with a single clarification turn providing substantial gains and recursive multi-turn execution offering modest additional improvements in select settings.\footnote{Our dataset and code can be found at \href{https://github.com/zwverse/clarify-or-answer}{https://github.com/zwverse/clarify-or-answer}}
\end{abstract}

\vspace{-1mm}
\section{Introduction}
\begin{figure*}[!t]
    \vspace{-3.2cm}
    \centering
    \includegraphics[width=\linewidth]{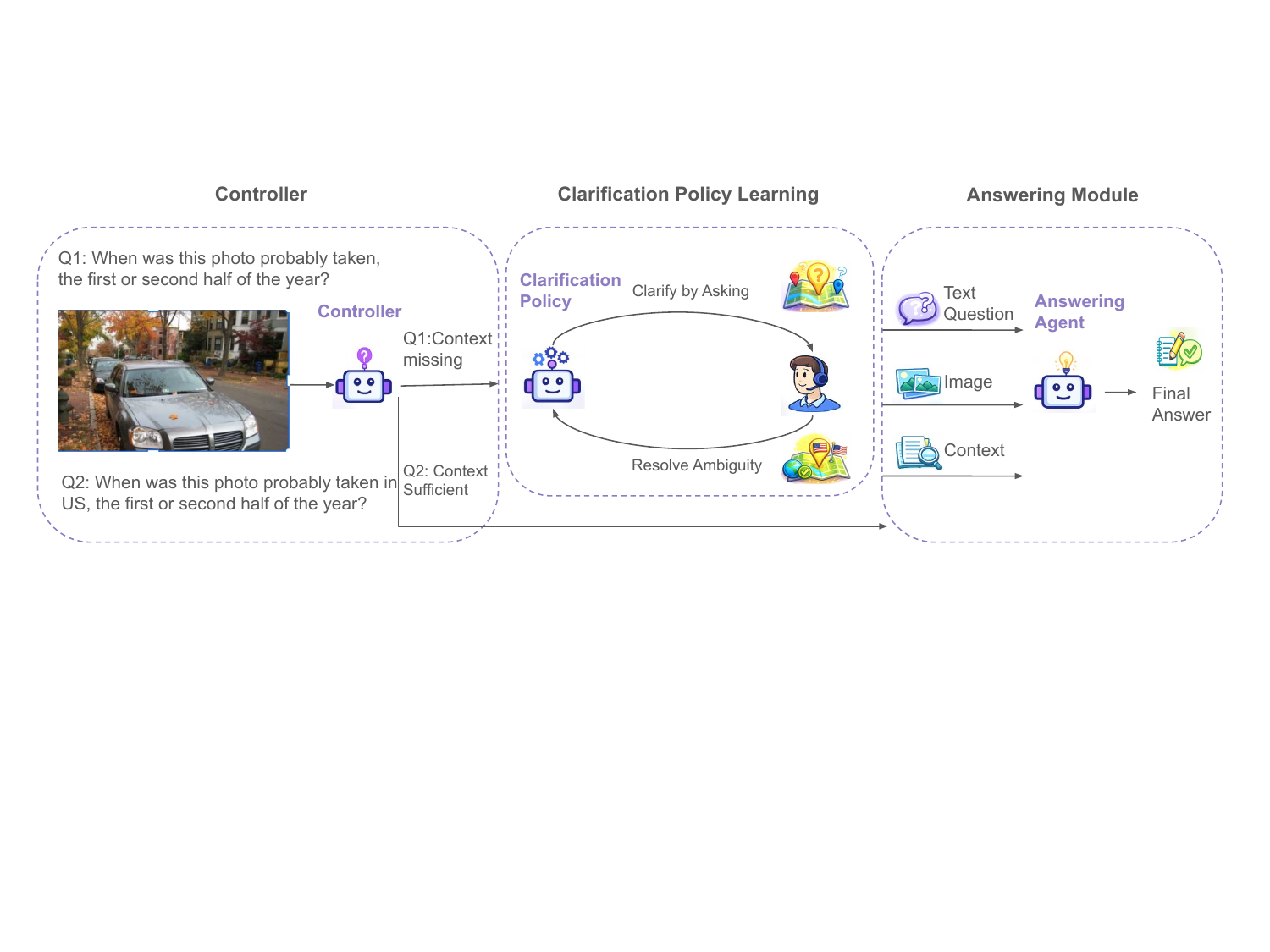}
    \vspace{-4.8cm}
    \caption{System overview of Clarify-or-Answer (CoA). The controller decides whether a visual question requires clarification. Ambiguous queries with missing context trigger a clarification step (clarification policy learning), while context-sufficient queries are directly forwarded to the VQA answering module, which produces the final answer.}
    \label{fig:system}
\end{figure*}

\vspace{-1mm}
Visual Question Answering (VQA) asks a model to answer a natural language question given an image \citep{agrawal_2016_vqa}. 
Real-world VQA is an interactive decision problem: an agent must decide whether it can answer reliably from the image alone, or whether it should first \emph{seek missing context} that is not visually observable \citep{luo2024codis}. When the input is underspecified, directly answering tends to produce confident but incorrect predictions. For example, in Figure~\ref{fig:system}, the question \emph{When was this photo taken, the first or second half of the year?} cannot be answered without knowing the \emph{location} of the photographer. A reliable VQA agent should therefore adapt its behavior: it should answer immediately when the VQA is sufficiently specified, and otherwise ask a targeted clarification question to obtain the missing information before answering.

Recent research has examined multiple sources of ambiguity in VQA, including perceptual ambiguity and visual illusions \citep{guan_2023_hallusionbench}, referential and focus-based uncertainty \citep{chen_2025_acknowledging}, multilingual and context-dependent ambiguity \citep{luo2024codis, wang_2025_mucar, wen-etal-2024-characterizing}, and robustness to rephrasings or visual corruption \citep{shah_2019_cycleconsistency, farhan_2024_visual}. Interactive formulations, e.g., ClearVQA, encourage asking clarification questions before answering when inputs are ambiguous \citep{jian_2025_teaching}. However, in prior work, ambiguity typically resides in the question phrasing or image content itself, not in the combination of both. This distinction inspires our focus on a complementary setting: underspecified VQA pairs, where the image and question individually are clear, but together yield multiple potentially valid answers depending on external context.

In response, we propose CoA (Clarify-or-Answer), an agentic framework that explicitly separates the decision of whether to answer from clarification question generation. CoA consists of three components: (i) a \emph{controller} that decides whether the agent should answer or ask for clarification, (ii) a \emph{clarification policy} that generates a single targeted clarification question when needed, and (iii) an \emph{answering model} that produces the final answer using the available context (Figure~\ref{fig:system}). 

We further study how to train the clarification policy of CoA to ask \emph{effective} questions rather than generic restatements. We curate \dataset, a dataset of 266 ambiguous image-question pairs derived from CODIS \citep{luo2024codis} and supplemented with manually sourced examples, each annotated with a human-verified, open-ended clarification question. To train and evaluate the controller, we construct a contrast set by augmenting each ambiguous instance with context-completed and irrelevant-context variants, yielding cases where clarification is necessary and where it is not. For learning to ask clarification questions, we introduce GRPO-CR, a reinforcement learning approach based on GRPO \citep{shao2024deepseekmath} that optimizes a multi-signal reward capturing question format, relevance to missing context, novelty, similarity to human clarifications, and ambiguity resolution potential. 

We evaluate CoA at both the module level (controller decisions; clarification quality) and the system level (overall VQA accuracy), comparing against prompting and supervised baselines. While CoA can operate as a single-step interaction, it also supports a recursive multi-turn paradigm where the agent iteratively asks one or more clarification questions to enrich the context until the ambiguity is resolved. Our results demonstrate that CoA outperforms prompting and supervised finetuning baselines; a single turn provides substantial VQA accuracy gains on ambiguous questions, and multi-turn execution offers small additional boosts to accuracy in most settings. 

In summary, our contributions are:
\begin{itemize}[topsep=0pt, itemsep=0pt, leftmargin=14pt]
    \item We propose CoA, an ask-or-answer agent that separates \emph{when to ask} (controller) from \emph{what to ask} (clarification policy), enabling adaptive decision making between clarifying and answering.
    \item We define the task of VQA clarification question generation in cases of context underspecification. We introduce \dataset, a dataset of 266 ambiguous VQA instances with human-verified open-ended clarifications and an additional 266 non-ambiguous contrast instances for training and evaluating ask-or-answer decisions.
    \item We develop GRPO-CR, a GRPO-based reinforcement learning method that improves clarification quality. CoA with GRPO-CR consistently outperforms prompting and finetuned baselines in controller logic, clarification policy execution, and final VQA accuracy across all tested datasets.
\end{itemize}

\vspace{-1mm}
\section{Related Work}
\vspace{-1mm}
\paragraph{Visual Ambiguity}
Visual ambiguity may arise from incomplete visual cues or distracting noise in a scene \citep{denison_2018_humans}. Much recent work on ambiguity detection focuses on benchmark creation and assessing how well off-the-shelf vision-language models (VLMs) handle different types of ambiguity~\citep{yue2024mmmu, yao2025mmmg, yang2025escaping}.
Some studies examine ambiguities caused by optical illusions \citep{cui_2023_holistic,fu_2023_a,guan_2023_hallusionbench}. Datasets like CODIS \citep{luo2024codis} evaluate whether models interpret images correctly across different contextual settings, while MUCAR \citep{wang_2025_mucar} introduces dual-ambiguity scenarios spanning both textual and visual dimensions. Audits of VQA benchmarks further demonstrate how underspecified temporal, spatial, or attribute scope introduces recurring visual ambiguity \cite{Qian2026Identifying}. Recently, ClearVQA \citep{jian_2025_teaching} proposes an interactive dataset where models must ask clarification questions when faced with ambiguous visual questions, and introduces a DPO-based training method to promote clarification-seeking behavior.

Our work is complementary to and differs from ClearVQA \citep{jian_2025_teaching} in several ways. In our setting, neither the image nor text is inherently ambiguous; instead, the VQA pair is underspecified, allowing multiple reasonable answers depending on external context. Thus, while both tasks involve clarification question generation, the underlying use cases are distinct. 
Further, ClearVQA generates Yes/No clarification questions, whereas our work focuses on open-ended clarification. Methodologically, we adopt GRPO reinforcement learning, in contrast to the DPO framework \citep{rafailov_2023_direct} used in ClearVQA.

\paragraph{Ambiguity Resolution through Question Generation}
Beyond benchmark construction, a growing body of research explores how models proactively ask clarification questions to resolve uncertainty. For text-based tasks, clarification question generation has been studied in dialogue and QA settings, where models are trained to request missing information rather than produce incomplete or incorrect answers \citep{rao_2018_learning, aliannejadi_2019_asking, wen-etal-2025-know}. Recent language model work has extended this idea, using preference optimization or reinforcement learning to encourage clarification seeking in conversational agents \citep{jq_2024_modeling,li_2025_alfa,zhang_2023_clarify, wen2023infovisdial}. 

However, work in multimodal domains remains limited.
While ClearVQA \citep{jian_2025_teaching} pioneers clarification question generation in VQA and recent approaches explore candidate answer generation under uncertainty \citep{Wang2026VQA}, our work extends this prior work on ambiguity resolution to context-dependent and ambiguous VQA pairs, providing a dataset and training method to enhance performance on this task.

\vspace{-1mm}
\section{Methodology}
\vspace{-1mm}
\subsection{Problem Setup}
\vspace{-1mm}
\label{sec:problem_setup}

We study \emph{context-dependent} VQA, where answering a question $Q_0$ given an image $V$ may require external context $c$ that is not visually observable. Following prior work on contextual ambiguity \citep{luo2024codis}, we consider five common underspecification types: (1) location and orientation, (2) temporal information, (3) cultural background, (4) attributes, and (5) relationships. Formally, the agent must decide whether the available context $\mathcal{X}_t$ (where $\mathcal{X}_0 = \{Q_0, V\}$) is sufficient to produce a correct answer $A$, or whether it must first elicit missing information. We model this as a decision problem where the agent selects an action $u \in \{\textsc{Answer}, \textsc{Clarify}\}$. If the agent chooses \textsc{Clarify}, it generates a clarification question $C_t$ to retrieve a response $R_t$, thereby augmenting the context $\mathcal{X}_{t+1} = \{\mathcal{X}_t, C_t, R_t\}$.

\subsection{CoA: Clarify-or-Answer Agent}
\vspace{-1mm}
\label{sec:coa_arch}

We propose CoA to resolve contextual ambiguity in this setting. CoA consists of three components:
\begin{itemize}[itemsep=0pt, topsep=1pt, leftmargin=16pt]
    \item \textbf{Controller} $\pi_{\text{ctrl}}(u \mid \mathcal{X}_t)$: Selects the agent action $u \in \{\textsc{Answer}, \textsc{Clarify}\}$
    \item \textbf{Clarification policy} $\pi_{\text{clr}}(C_t \mid \mathcal{X}_t)$: Generates a targeted clarification question $C_t$ when $u=\textsc{Clarify}$
    \item \textbf{Answering model} $\pi_{\text{ans}}(A \mid \mathcal{X}_t)$: Outputs the final answer $A$ conditioned on the accumulated context $\mathcal{X}_t$
\end{itemize}

\noindent At inference time, CoA evaluates the current context $\mathcal{X}_t$ using the controller. If $u = \textsc{Answer}$, the agent provides the final answer $A$. If $u = \textsc{Clarify}$, it generates $C_t$, receives $R_t$, and updates the context to $\mathcal{X}_{t+1}$.  Decoupling the controller and policy avoids conflicting supervision: the controller requires training on mixed contrast sets to judge ambiguity, while the policy is optimized via RL strictly on missing-context instances.

\subsection{Controller Learning Objective}
\vspace{-1mm}
\label{sec:controller_objective}

The controller is trained to predict whether missing context is required to answer reliably. We formulate this as binary classification over $u \in \{\textsc{Answer}, \textsc{Clarify}\}$ and optimize the standard cross-entropy loss $\mathcal{L}_{\text{ctrl}} = -\log \pi_{\text{ctrl}}(u^\ast \mid \mathcal{X}_0)$, where $u^\ast$ is the ground-truth ask-or-answer label for the initial state.

\subsection{Clarification Policy Learning}
\vspace{-1mm}
\label{sec:clarification_learning}

For ambiguous inputs, the clarification policy is trained to generate a question $C_t$ that effectively elicits the missing contextual factor $c$.

\begin{figure*}[t!]
    \vspace{-4mm}
    \centering
    \includegraphics[width=0.94\linewidth]{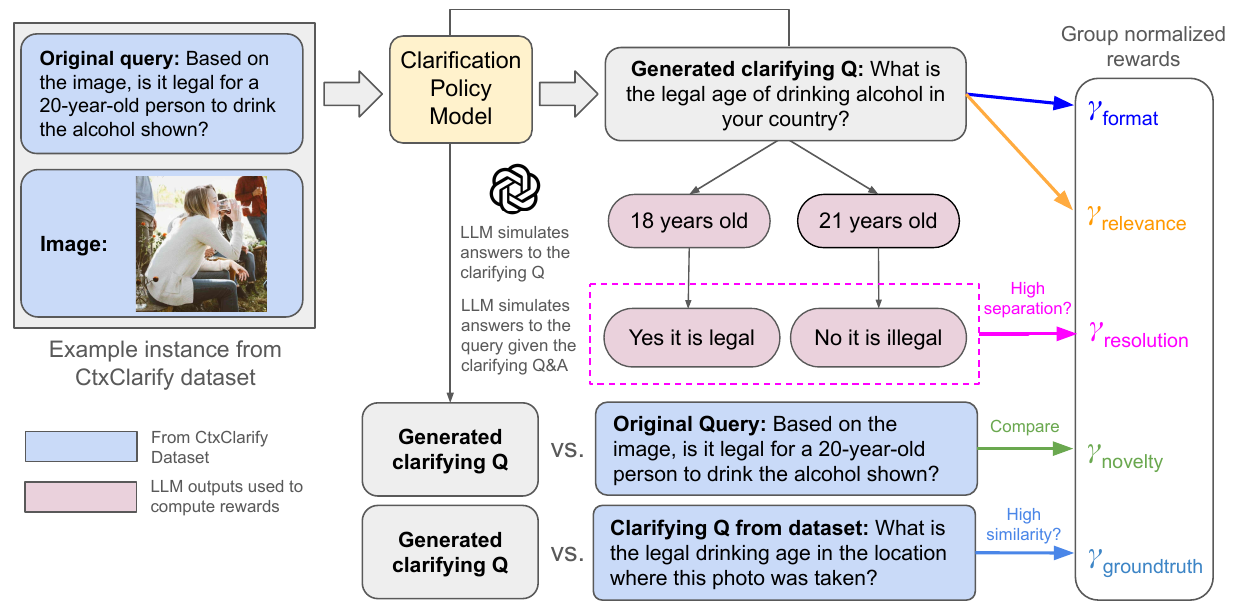}
    \caption{Training pipeline for our method GRPO-CR. The clarification policy model generates clarification questions from image-query pairs, which are evaluated by multi-dimensional rewards and updated via GRPO.} 
    \label{fig:training_pipeline}
    \vspace{-1mm}
\end{figure*}

\paragraph{Reinforcement learning with GRPO-CR.}
We introduce GRPO-CR (Clarification Reasoning) (see Figure~\ref{fig:training_pipeline}), a reinforcement learning approach built on Group Relative Policy Optimization (GRPO) \citep{shao2024deepseekmath}. For each input $\mathcal{X}_t$, the policy samples a group of $N$ candidate clarifications $\{C_t^{(i)}\}_{i=1}^N$ and receives a scalar reward $R(C_t^{(i)})$. We compute within-group relative advantages as:
\vspace{-1mm}
\begin{equation*}
\hat{A}_i = R(C_t^{(i)}) - \frac{1}{N}\sum_{j=1}^{N} R(C_t^{(j)})
\end{equation*}
\noindent and optimize the policy loss:
\vspace{-1mm}
\begin{equation*}
\mathcal{L}_{\text{GRPO-CR}} =
-\mathbb{E}_{i}\left[
    \log \pi_{\text{clr}}
    \left(C_t^{(i)} \mid \mathcal{X}_t\right)
    \cdot \hat{A}_i
\right]
\end{equation*}

\vspace{-2mm}
\paragraph{Reward Components}
Our reward design builds on prior work showing that effective clarification requires balancing structural validity and informativeness. For example, ALFA demonstrates the value of rewards capturing both format compliance and utility \citep{li_2025_alfa}, while GRIT highlights the need for explicit rewards in multimodal reasoning \citep{fan_2025_grit}. In contrast to text-only or general reasoning settings, we introduce a tailored set of multimodal rewards designed for context underspecification in VQA.

The total reward $R(C_t)$ is a weighted sum of individual rewards $R(C_t) = \sum_{k} \lambda_k R_k(C_t)$ with $\lambda_k = 1$. Individual reward components are determined based on relative importance and a parameter sweep over reward coefficients (Appendix~\ref{app:appendix_reward_sweep}).

\begin{itemize}[topsep=1pt, leftmargin=16pt, itemsep=0pt]
    \item \textbf{Format reward ($\gamma_{\text{format}}$)}: The format reward encourages outputs that are well-formed, concise clarification questions. Assigns $+0.5$ if $C_t$ ends with a question mark and is under 50 words; otherwise $-0.5$.

    \item \textbf{Focused relevance reward ($\gamma_{\text{relevance}}$)}: This reward encourages clarification questions that explicitly target the missing context. We define a series of lexical patterns (e.g., \emph{what}, \emph{which}) and keywords associated with different categories of questions (e.g., \emph{country}, \emph{city} for cultural ambiguity) (full descriptions and lists in Appendix~\ref{sec:relevance_reward_details}). We assign higher scores to questions that (i) follow clarification-oriented patterns and (ii) include any keywords associated with the type of ambiguity. $C_t$ satisfying both conditions receive $+0.3$, those satisfying only one receive $+0.1$, and others receive $-0.1$.
    \item \textbf{Ambiguity resolution reward ($\gamma_{\text{resolution}}$)}: Assigns $+0.5$ if the response $R_t$ leads to a different final answer $A$, indicating the ambiguity was resolved, and $-0.3$ otherwise.
    \item \textbf{Novelty reward ($R_{\text{novelty}}$)}: This reward penalizes trivial rephrasings of the original question $Q_0$ by encouraging semantic diversity. We compute the Jaccard similarity $J$ between the clarification question $C_t$ and $Q_0$. If $J > 0.8$, the reward is $-0.3$; if $0.8 \geq J \geq 0.6$, the reward is $-0.1$; if $J < 0.6$, the reward is $+0.3$.
    \item \textbf{Ground-truth check reward ($\gamma_{\text{groundtruth}}$)}: This reward measures alignment between the generated clarification question and human-authored reference clarifications. It uses GPT-4o~\citep{Hurst2024GPT4oSC} to assign a similarity score in $[0,1]$ between $C_t$ and a human-authored reference.
\end{itemize}

We further validate our reward design via reward ablation studies to evaluate robustness. Details are reported in the Appendix~\ref{app:appendix_reward_ablation}.

\subsection{End-to-End CoA Execution}
\label{sec:end2end_exec}The CoA framework is evaluated under two execution settings:

\textbf{Single-turn:}~The model performs a one-step decision on $\mathcal{X}_0$. It either generates a direct answer $A$ or issues exactly one clarification question $C_1$. Upon receiving $R_1$, the framework produces the final answer $A$ based on the augmented context $\mathcal{X}_1 = \{Q_0, V, C_1, R_1\}$.

\textbf{Multi-turn:}~The clarification process is treated as a recursive loop. At each turn $t$, the model evaluates $\mathcal{X}_t$. If it selects \textsc{Clarify}, it generates $C_t$ and receives $R_t$, forming $\mathcal{X}_{t+1} = \{\mathcal{X}_t, C_t, R_t\}$. This context is fed back into the CoA framework to re-evaluate the evidence. If the model further selects \textsc{Clarify}, new $C_t$ and $R_t$ are appended to the context. This continues until the model selects \textsc{Answer} or reaches a maximum turn limit $T$. If the turn limit $T$ is reached, the current augmented context is used to produce a final answer.

\vspace{-1mm}
\section{\dataset Dataset}
\vspace{-1mm}
\label{sec:dataset}
To validate CoA and GRPO-CR on underspecified VQA pairs, we introduce \dataset, a curated dataset of context-dependent VQA pairs annotated with open-ended clarification questions, and a corresponding contrast set with non-ambiguous VQA pairs.

\paragraph{Source Dataset}
Our dataset is derived from  CODIS \citep{luo2024codis}, which contains
222 images paired with 395 questions. CODIS questions exhibit diverse forms of contextual ambiguity, including categories such as \emph{location and orientation}, \emph{temporal information}, \emph{cultural background}, \emph{attributes}, and \emph{relationships}. While this makes CODIS well suited for studying ambiguity, not all question-image pairs are appropriate for learning targeted clarification behavior, as some questions are vague or require resolving multiple missing factors.

\paragraph{Dataset Curation and Clarification Annotation}
From the original CODIS dataset, we curate a subset of 266 high-quality VQA items that each depend on a single missing contextual factor. Approximately 30\% of CODIS pairs are discarded due to poor visual grounding or compound ambiguities. To improve clarity, we further replace about 10\% CODIS images with carefully matched ambiguous images collected from the Internet. These replacements preserve the original contextual dependencies while providing stronger visual grounding.

For each retained item, we annotate one clarification question that explicitly targets missing context required to answer the question correctly. We first use GPT-4o to generate candidate clarification questions given the image, question, and known ambiguous contexts, and then manually review the outputs for acceptability. During review, annotators either accept the generated question without change or refine it to improve precision. This curated subset of CODIS along with our annotated clarification questions are included in \dataset.

\paragraph{Contrast Dataset Construction}
For each ambiguous item from before, we construct two variants: (i) a fully specified version where the missing relevant context is added into the question, yielding a question that no longer requires clarification, and (ii) a version augmented with additional but irrelevant contextual information that does not resolve the ambiguity and still requires clarification. This procedure results in a mixture of questions that either do or do not require clarification that we use to train our Controller. Additional details on ambiguity categories, mixed-dataset construction rules, and illustrative examples are provided in Appendix~\ref{sec:dataset_appendix}.

\paragraph{Final Dataset Statistics} 
\dataset and its contrast set contain 532 VQA instances split 50/50 between ambiguous and non-ambiguous contrast items across 70\% train, 15\% validation, and 15\% test splits. It spans five contextual ambiguity categories: \emph{location and orientation}, \emph{temporal information}, \emph{cultural background}, \emph{attributes}, and \emph{relationships}. See Appendix~\ref{app:dataset_breakdown} for  full breakdown.

\vspace{-2mm}
\section{Experimental Setup}
\vspace{-1mm}
\subsection{Datasets, Models \& Baselines}
\vspace{-1mm}
\label{sec:exp_setup}
\paragraph{Datasets}
\label{sec:exp_datasets}  We use \dataset (abbreviated CtxClarify) for training and evaluation of CoA. To assess generalization beyond \dataset, we also evaluate on VizWiz \citep{chen_2025_acknowledging} and ClearVQA \citep{jian_2025_teaching}, using their original train/dev/test splits. Both VizWiz and ClearVQA contain different types of visual ambiguity compared to \dataset.

\vspace{-1mm}
\paragraph{Models} We adopt three VLMs of varying scales: InternVL3-2B-Instruct~\citep{zhu_2025_internvl3}, Qwen2.5-VL-7B-Instruct~\citep{qwen_2024_qwen25} and Qwen2.5-VL-3B-Instruct~\citep{qwen_2024_qwen25} as backbones for the controller, clarification policy, and answering models in our framework (referred to as IVL-2B, Qw-3B, and Qw-7B, respectively). Additionally, to provide reference points against state-of-the-art commercial systems, we evaluate closed-weight frontier models GPT-4o \citep{OpenAI2024GPT} and Gemini-2.5-Flash \citep{Comanici2025Gemini} under a zero-shot prompting configuration.

\vspace{-1mm}
\paragraph{Baselines} \label{sec:exp_baselines} We compare CoA against baselines at both the system and module level. For the system-level baseline, we use (i) \emph{prompting}, where the backbone model directly decides whether to answer or ask a clarification question, (ii) \emph{always ask (AA$_{GRPO}$)}, where the GRPO-trained clarification model bypasses the controller and always produces a clarification question, (iii) \emph{in-context learning (ICL)} using 4 few-shot examples (Appendix~\ref{app:icl_setup}), and (iv) an \emph{oracle setting} ($\text{CoA}_{\text{Oracle}}$), where the controller is assumed to be 100\% accurate at providing ground-truth ask-or-answer labels, isolating downstream clarification policy performance.

For module-level baselines: (i) Controller module: we \emph{prompt} the backbone model to directly classify whether a question is ambiguous or not. (ii) Clarification module: we compare against \emph{prompting}, where the backbone model is directly prompted to generate a clarification question, and \emph{supervised finetuning (SFT)}, where the backbone model is trained on \dataset to generate clarification questions.

\vspace{-1mm}
\subsection{Implementation Details}
\vspace{-1mm}
\label{sec:exp_training}

\paragraph{Controller training.}
We train controllers on \dataset (over the train split of ambiguous questions and the contrast set) with cross-entropy over $\{\textsc{Answer},\textsc{Clarify}\}$. We implement supervised finetuning using MS-SWIFT \citep{zhao_2024_swifta}. Prompting baselines use a fixed decision prompt (Appendix~\ref{app:prompt_templates}).

\vspace{-1mm}
\paragraph{Clarification policy training.}
We train only on the train split of the ambiguous subset of \dataset. The policy described in \S\ref{sec:clarification_learning} is optimized using GRPO to refine clarification strategies through relative reward signals.

\input{tables/ctxclarify_results_table}
\vspace{-1mm}
\paragraph{End-to-end CoA execution.}
In the end-to-end setting, CoA either answers directly when the controller selects \textsc{Answer} or generates a question when the controller selects \textsc{Clarify}. When a clarification question is generated, we use GPT-4o to simulate a plausible user response to the question conditioned on the image (following prior work such as~\citet{jian_2025_teaching}), and condition the answering model on all of the above when producing the final answer. This avoids reliance on oracle context while allowing us to assess whether the learned clarification strategies translate to improvements in downstream VQA accuracy.

All modules use the same backbone model and follow identical training and inference protocols to ensure fair comparisons. 
Further implementation details are in Appendix~\ref{app:training_details}.

\subsection{Evaluation Metrics}
\vspace{-1mm}
\paragraph{VQA accuracy} We measure final answer correctness by reporting standard VQA accuracy \citep{agrawal_2016_vqa}. We evaluate VQA accuracy by comparing the agent's final predictions against ground-truth answers. We employ GPT-4o~\citep{Hurst2024GPT4oSC} as an automated judge to assign a binary score ($1$ for correct, $0$ for incorrect) to each VQA pair.

We compute accuracy in both single-turn and multi-turn settings as described previously. Once all questions in the dataset have received a final prediction, either through early resolution or by reaching the maximum turn limit $T$. The overall VQA accuracy is calculated across the full test set across 3 independent runs, reporting 95\% confidence intervals.

\paragraph{Component-wise evaluation} We also report performance for separate components of CoA:
\vspace{-2mm}
\begin{itemize}[topsep=0pt, leftmargin=12pt, itemsep=0pt]
\item{\textbf{Controller performance:}} We report accuracy, precision, recall, and F1 for whether the controller correctly predicts \textsc{Clarify} versus \textsc{Answer} over the mixed ambiguous and contrast sets.
\item{\textbf{Clarification quality:}} Beyond VQA accuracy, we also evaluate clarification questions using the Ambiguity Resolution Reward metric introduced in our reward function $\gamma_{\text{resolution}}$ estimating whether answering the clarification question is likely to change the final VQA outcome. We additionally validate this reward metric with human judgments. We treat $\gamma_{\text{resolution}}$ as a diagnostic proxy for ambiguity resolution during policy training, relying on independent VQA accuracy as our primary metric to prevent reward hacking.
\end{itemize}
\paragraph{Human validation} To ensure metric reliability, the first authors also manually annotate a sample of the test data across models for both VQA accuracy and the quality of generated clarification questions. We randomly sample and annotate 20\% of the test dataset for VQA accuracy using the same binary score standard given to GPT-4o, and 30\% of the \dataset test set to evaluate whether the generated clarification question resolves ambiguity from the original question. Human validation instructions and operationalization details can be found in~Appendix~\ref{app:human_validation}.

\vspace{-1mm}
\section{Results \& Discussion}
\vspace{-1mm}
\subsection{VQA Accuracy}
\vspace{-2mm}
\label{sec:vqa_accuracy}\
\paragraph{Overall Performance Enhancement} 
CoA improves overall VQA accuracy for the tested open-weight models over prompting and other baselines on \dataset (Table~\ref{tab:model_results}). Finetuned smaller models narrow the gap to frontier models (GPT-4o, Gemini-2.5-Flash), especially in categories such as \emph{Cultural Background} or \emph{Temporal Information}. While SFT provides consistent gains, GRPO-CR leads to larger improvements on average, outperforming zero-shot prompting, Always-Ask ($\text{AA}_{\text{GRPO}}$), and ICL baselines.

\input{tables/ood_results_table}
\paragraph{OOD Generalization} Performance gains generalize to OOD datasets ClearVQA and VizWiz (Table~\ref{tab:vqa_ood_results}). Across all three base models, CoA with GRPO yields consistent improvements to final VQA accuracy, even though both OOD datasets consist of different types of visual/textual ambiguity than those included in \dataset.

\input{figures/multiturn_vqa_accuracy}
\paragraph{Effect of Multi-turn Clarification}
Figure~\ref{fig:multi_turn_vqa} shows VQA accuracy as a function of the number of clarification turns. Most methods benefit from additional clarification turns, but the additional gains are marginal, and in a few settings actually lead to reduced overall accuracy. This suggests that the most critical ambiguities can be resolved early, and that extended interaction mostly leads to consistent or minor degradations in context.

\paragraph{Error Analysis} We examine model decisions, clarification questions, and answers to understand how CoA$_{SFT+GRPO}$ behaves across ambiguous and unambiguous VQA scenarios compared to baselines. Figure~\ref{fig:case_studies_main_paper} shows select examples (more in Appendix~\ref{app:case_study}).
When questions are underspecified, CoA$_{SFT+GRPO}$ identifies missing contextual factors and generates targeted clarification questions that either fully resolve the ambiguity or narrow it to support a more reliable answer. In contrast, baseline prompting results in answers based on unwarranted inferences from partial visual cues. When questions are unambiguous, CoA$_{SFT+GRPO}$ avoids unnecessary clarification and demonstrates an ability to answer directly, indicating that gains generally stem from selective and effective clarification rather than always asking.
\input{figures/case_studies_main}
\subsection{Module-level Performance}
\label{sec:ambiguity_detection}

\input{tables/module_level_results}
\paragraph{Ambiguity Detection and Controller Generalization} Table~\ref{tab:controller_perf} shows training the CoA controller with SFT consistently improves ambiguity detection over zero-shot prompting on \dataset and generalizes to the ClearVQA setting. The results show that the controller is much better at catching cases where a question is unclear, whereas the basic prompting method often misses them entirely. This is especially true for smaller models like IVL-2B, which struggle to trigger any clarifications without this specific training. Most importantly, these gains hold up even on the OOD ClearVQA data (we do not evaluate on VizWiz because it only contains ambiguous instances without a contrast set). This suggests that the model does not just memorize specific examples, but has learned general patterns for when to ask for more information.

\paragraph{Controller Bottleneck Analysis.} To test how much controller errors cap end-to-end performance, we evaluate an \emph{Oracle Controller} ($\text{CoA}_{\text{Oracle}}$) with ground-truth decision labels. As shown in Tables~\ref{tab:model_results} and~\ref{tab:vqa_ood_results}, replacing the learned controller with an oracle provides only modest accuracy gains. This demonstrates that while ambiguity detection is non-trivial, the controller is not the main system bottleneck; downstream clarification quality and context integration play an equally critical role.

\paragraph{Clarification Question Quality Improvements} CoA$_{SFT+GRPO}$ improves clarification question quality over both prompting and supervised fine-tuning, and generalizes to OOD datasets.
Table~\ref{tab:clarification_policy_combined} reports ambiguity resolution scores across all datasets and model families. 

On \dataset, CoA$_{SFT+GRPO}$ achieves the best ambiguity resolution across all evaluated models, substantially improvingover both prompting and SFT. Gains are especially pronounced for Qwen-7B and InternVL-2B. These improvements also transfer to ClearVQA and Vizwiz, where CoA$_{SFT+GRPO}$ consistently yields the highest scores across models, showing that the learned clarification strategy does not overfit to \dataset and remains effective under distribution shift.

\subsection{Human Validation}
\paragraph{Human Evaluation of Generated Clarification Questions} We conduct a human annotation (Appendix~\ref{app:human_validation}) across three models on \dataset in Table~\ref{tab:clarification_policy_human}. We found an 84\% percentage agreement between human and GPT-based evaluations, corresponding to a Cohen’s Kappa of 0.660. This indicates substantial agreement, confirming that the LLM judge is a reliable proxy for evaluating clarification quality.

\paragraph{Human Validation of LLM Judge} To ensure reliability of our automated evaluation, we conduct a human correlation study (Appendix~\ref{app:human_vqa_eval}) and calculate percent agreement ($PA$) between GPT-based scores and human judgments. Results demonstrate positive correlation ($PA \approx 82\%$). GPT evaluation is more conservative than human judgment. While humans use common sense to capture hidden context, the LLM is more literal and requires direct visual proof to mark an answer accurate. This shows that while our auto-eval metric introduces some error, it serves as a reasonable proxy for human perception on this judgment task.

\vspace{-2mm}
\section{Limitations \& Future Work}
\vspace{-1mm}
\label{sec:limitations}
\input{tables/human_eval_vqa}
\paragraph{Broader coverage and generalization.}
\dataset is derived from CODIS and focuses on single-factor underspecification. Extending coverage to more domains (e.g., diagrams, documents, embodied scenes), more diverse question styles, and more languages would improve external validity. Additionally, future datasets could explicitly annotate multiple plausible missing factors to study how clarification policies prioritize which question to ask first, and whether the agent can ask the most informative question under a one-question budget.

\paragraph{End-to-end optimization and richer objectives.}
CoA currently optimizes the controller and clarification policy, while treating the answering model as a fixed backbone conditioned on the original inputs (or augmented with the clarification response). Joint optimization of all components could yield further gains but raises credit-assignment and stability challenges. Another direction is to optimize for user-centric objectives such as helpfulness, succinctness, and trust calibration, in addition to final answer accuracy.

\paragraph{Role of GPT-4o and real-world limitations.}
We rely heavily on GPT-4o to draft initial dataset candidates, compute similarity rewards during RL training, simulate user responses during testing, and judge VQA answer accuracy. To ensure quality and reduce bias, the authors manually review and refine all dataset questions, and validate a subset of responses for both RL rewards and VQA accuracy. While simulated responses allow for benchmarking feasibility, real human input is naturally noisier. 
Naturally, testing with human participants remains an important next step.

\section{Conclusion}
\label{sec:conclusion}
We present CoA (Clarify-or-Answer), an ask-or-answer agentic framework for context-dependent VQA that separates \emph{when to ask} (controller) from \emph{what to ask} (clarification policy), and produces a final answer either directly or after incorporating a single clarification response. We introduce \dataset, a dataset comprising 266 ambiguous instances with human-verified, open-ended clarification questions and 266 non-ambiguous contrast instances with complete context, enabling training and evaluation of ask-or-answer decisions and clarification generation. To learn effective clarification behavior, we propose GRPO-CR, a GRPO reinforcement learning method with multi-signal rewards encouraging focused, ambiguity-resolving questions. Our framework supports both single-turn and recursive multi-turn execution. The first turn provides the most performance gains and the recursive multi-turn paradigm further boosts accuracy by iteratively enriching the context. Across module and system level evaluations, and different model families, CoA$_{SFT+GRPO}$ achieves the highest overall VQA accuracy compared to zero-shot prompting, ICL, $\text{AA}_{\text{GRPO}}$ and CoA$_{SFT}$, approaching the upper bound of an oracle controller ($\text{CoA}_{\text{Oracle}}$) and highlighting the value of explicit clarification for context-dependent visual and textual reasoning.

\section*{Acknowledgements}
\label{sec:acks}
This work is partially supported by gift funds from Google and the Allen Institute for AI.

\section*{Ethical Statement}
\label{sec:ethical_statement}
\textbf{Potential Risks.} 
This work studies clarification question generation for ambiguous VQA. While clarification can improve reliability, models may ask inappropriate or biased questions, increase user burden, or be misused to elicit sensitive information if deployed without safeguards. To mitigate these risks, our dataset excludes personally identifiable or sensitive content and focuses on benign contextual ambiguities. This work is intended for research use and should not be directly deployed in high stakes domains without additional safeguards and human oversight.

\textbf{Use of AI Assistants.} 
AI assistants were used to support paper writing, develop data-processing and evaluation scripts, simulate clarification responses, compute similarity-based rewards, and generate initial drafts of clarification questions that are subsequently reviewed and/or modified by the authors to ensure accuracy and clarity. All AI-generated content was reviewed and validated by the authors, who retain full responsibility for the dataset, methodology, and conclusions.

\bibliography{colm2026_conference}
\bibliographystyle{colm2026_conference}

\appendix

\section{GRPO Reward Function Parameter Sweep}
\label{app:appendix_reward_sweep}
We observe that clarification effectiveness remains relatively stable across moderate variations in reward scales, indicating that the proposed reward design is not highly sensitive to precise coefficient choices (Figure~\ref{tab:reward_sweep}). 

The picked configuration consistently performs near the top, supporting our empirical selection. Among the components, overly strong format rewards degrade performance, suggesting that enforcing structural correctness too aggressively can hinder informativeness, while weaker format rewards are sufficient as a soft constraint. Ambiguity and relevance rewards show stable behavior across settings, indicating that their contributions are important but not dependent on fine-grained tuning. Overall, these results suggest that the model benefits from a balanced combination of reward signals rather than precise optimization of individual coefficients.

\paragraph{Reward Scaling Rationale.}
We assign equal weights ($\lambda_k = 1.0$) across reward components to maintain a simple, un-tuned baseline and prevent overfitting to a specific reward configuration. Individual reward scales were bounded within a standard $[-1.0, +1.0]$ interval based on pilot experiments and established RL practice \citep{shao2024deepseekmath, fan_2025_grit}, ensuring that format compliance, relevance, and ambiguity resolution signals maintain comparable magnitudes during GRPO optimization.

\section{Details of Focused Relevance Reward}
\label{sec:relevance_reward_details}

To address the ambiguity categories and the scoring criteria for the Focused Relevance Reward ($\gamma_{\text{relevance}}$), we provide the specific keyword mappings and linguistic patterns used during training.

\paragraph{Keywords Per Ambiguity Type}
A clarification question $C_t$ satisfies the \textbf{Keywords} condition if it contains at least one token from the set associated with its specific ambiguity type. These categories and their respective keywords are defined in Table \ref{tab:keywords}.

\begin{table}[h]
\centering
\small
\begin{tabular}{lp{10cm}}
\toprule
\textbf{Category} & \textbf{Keywords} \\ \midrule
Cultural Background & country, city, state, region, location, place, area, nation, where, origin, cultural, tradition, custom, law, legal, illegal, rules, culture \\ \midrule
Temporal Information & before, after, prior, following, earlier, later, recently, soon, now, current, time, date, day, night, morning, afternoon, evening, today, tomorrow, yesterday \\ \midrule
Location \& Orientation & left, right, up, down, forward, backward, toward, away, direction, push, pull, open, close, above, below, enter, exit \\ \midrule
Attributes & hot, cold, on, off, high, low, big, small, large, short, long, heavy, light, temperature, speed, limit, safe, dangerous, bright, dark, clean, dirty, full, empty \\ \midrule
Relationships & who, which, person, man, woman, boy, girl, friend, family, colleague, partner, teacher, student, owner, driver, passenger, speaker, listener, helper \\ \bottomrule
\end{tabular}
\caption{Mapping of Ambiguity Categories to Targeted Keywords}
\label{tab:keywords}
\end{table}

\paragraph{Clarification Patterns}  The pattern-matching condition ensures that questions are structurally oriented toward information seeking rather than generic conversational fillers.

\begin{itemize} [noitemsep, topsep=0pt, leftmargin=12pt]
    \item \textbf{Specific Patterns (Satisfies Condition)}: These identify questions targeting specific properties, alternatives, or comparative states:
    \begin{itemize}
        \item Explicit Choice: e.g., \emph{"either"}, \emph{"neither"}, or the use of \emph{"A or B"}.
        \item Property Targeting: short wh-questions such as \emph{"which [type/kind/part]?"} or \emph{"what [version/value]?"}.
        \item Reference Resolution: e.g., \emph{"which of them?"} or \emph{"what purpose?"}.
        \item Comparative Cues: e.g., \emph{"earlier"}, \emph{"smaller"}, \emph{"higher"}.
    \end{itemize}
    \item \textbf{Vague Patterns (Fails Condition)}: These identify generic or non-specific requests that fail the structural relevance test:
    \begin{itemize}
        \item Generic Requests: e.g., \emph{"tell me more"}, \emph{"elaborate"}, \emph{"give me details"}.
        \item Confusion Markers: e.g., \emph{"I don't understand"}, \emph{"not sure"}.
        \item Indefinite Pronouns: e.g., \emph{"anything"}, \emph{"something"}.
        \item Bare Why: a standalone \emph{"why?"} without a following subject or object.
    \end{itemize}
\end{itemize}

\paragraph{Reward Calculation}
The final score is assigned as follows:
\begin{equation}
\gamma_{\text{relevance}} = 
\begin{cases} 
+0.3 & \text{if both keyword and specific Pattern match} \\
+0.1 & \text{if only one condition matches} \\
-0.1 & \text{if neither condition matches}
\end{cases}
\end{equation}

\section{GRPO Reward Ablation}
\label{app:appendix_reward_ablation}
To evaluate the contribution of each reward component in GRPO-CR, we conduct an ablation study by selectively removing individual rewards while keeping the remaining components unchanged. We report clarification effectiveness on the CtxClarify dataset using InternVL3-2B.

\begin{table}[t!]
\centering
\small
\begin{tabular}{l l l l c}
\toprule
\textbf{Dataset} & \textbf{Model} & \textbf{Removed} & \textbf{Kept Rewards} & \textbf{Effectiveness} \\
\midrule
CtxClarify (Auto) & InternVL3-2B & None & GT + F + R + N + A & 0.248 $\pm$ 0.010 \\
CtxClarify (Auto) & InternVL3-2B & A & GT + F + R + N & 0.145 $\pm$ 0.101 \\
CtxClarify (Auto) & InternVL3-2B & N & GT + F + R + A & 0.107 $\pm$ 0.006 \\
CtxClarify (Auto) & InternVL3-2B & R & GT + F + N + A & 0.126 $\pm$ 0.003 \\
CtxClarify (Auto) & InternVL3-2B & A + N + R & GT + F & 0.101 $\pm$ 0.008 \\
\bottomrule
\end{tabular}
\caption{Ablation of GRPO-CR reward components. GT = Ground Truth similarity, F = Format, R = Relevance, N = Novelty, A = Ambiguity.}
\label{tab:reward_ablation}
\end{table}

\paragraph{Overall impact.}
As shown in Table~\ref{tab:reward_ablation}, removing any individual reward component leads to a substantial drop in clarification effectiveness, confirming that the reward signals are complementary and jointly necessary for high quality clarification generation.

\paragraph{Ambiguity reward.}
Removing the ambiguity reward results in a large performance drop and significantly increased variance. This suggests that explicitly incentivizing ambiguity resolution is critical for stabilizing training and guiding the model toward asking clarifying questions rather than prematurely answering.

\paragraph{Novelty reward.}
The largest degradation is observed when removing the novelty reward, indicating that without encouraging diverse and non-trivial questions, the model tends to generate generic or redundant clarifications that fail to resolve uncertainty effectively.

\paragraph{Relevance reward.}
Removing the relevance reward also reduces performance, highlighting its role in ensuring that clarification questions remain grounded in the visual and textual context rather than drifting into unrelated or overly broad queries.

\paragraph{Minimal reward setting.}
When only Ground Truth similarity and Format rewards are retained, performance drops further, demonstrating that structural correctness alone is insufficient; semantic rewards (ambiguity, novelty, relevance) are essential for meaningful clarification.

\paragraph{Takeaway.}
These results validate our joint reward design in GRPO-CR. Each component addresses a distinct failure mode and their combination is necessary to achieve robust clarification behavior.

\input{tables/appendix_case_studies}

\section{Dataset Details}
\label{sec:dataset_appendix}
\subsection{Data Split}
Table~\ref{tab:dataset_split} summarizes the data splits for the ambiguous subset of \dataset. The non-ambiguous subset follows the same split ratio.

\begin{table}[]
\centering
\small
\begin{tabular}{lcc}
\toprule
\textbf{Split} & \textbf{\# Samples} & \textbf{Percentage} \\
\midrule
Training   & 186 & 70\% \\
Validation &  40 & 15\% \\
Test       &  40 & 15\% \\
\midrule
\textbf{Total} & 266 & 100\% \\
\bottomrule
\end{tabular}
\caption{Dataset split of \dataset for the ambiguous subset. The non-ambiguous subset uses the same split ratio and preserves the corresponding questions in each split.}
\label{tab:dataset_split}
\end{table}
\subsection{Ambiguity Types}
Each ambiguous example in \dataset is associated with a single missing contextual factor. Following the categorization introduced in the original CODIS dataset \citep{luo2024codis}, we consider five types of ambiguity:
\begin{itemize}[noitemsep, topsep=0pt, leftmargin=18pt]
    \item Location and orientation: the answer depends on geographic location or spatial orientation (e.g., country, hemisphere).
    \item Temporal information: the answer depends on time-related context (e.g., season, historical period).
    \item Cultural background: the answer depends on cultural or social norms.
    \item Attributes: the answer depends on properties of entities not visually specified (e.g., age, status).
    \item Relationships: the answer depends on social or interpersonal relationships.
\end{itemize}

Each ambiguous example is curated to isolate exactly one ambiguity type.

\subsection{Ambiguity Category Breakdown and Annotation Setup}
\label{app:dataset_breakdown}

Table~\ref{tab:category_breakdown} provides a breakdown of \dataset across the five contextual ambiguity categories.

\begin{table}[h]
\centering
\small
\begin{tabular}{cccccc}
\toprule
\textbf{Attributes} & \textbf{Cultural Backgr.} & \textbf{Location \& Orient.} & \textbf{Relationships} & \textbf{Temporal Info.} & \textbf{Total} \\
\midrule
116 & 86 & 114 & 84 & 132 & 532 \\
\bottomrule
\end{tabular}
\caption{Distribution of instances across contextual ambiguity categories in \dataset.}
\label{tab:category_breakdown}
\end{table}

\paragraph{Annotator Qualifications and Procedure.}
Dataset curation and manual verification were conducted by the paper's primary authors (graduate researchers in Computer Science and NLP). Candidate clarification questions generated during preprocessing were manually reviewed, edited, or re-annotated to ensure precise visual grounding and single-factor targeting. Disagreements were resolved through consensus discussion among the authors.

\paragraph{Balanced Dataset Design.}
We intentionally construct a balanced $1:1$ mixture of ambiguous and context-complete instances in our contrast set. This setup provides a clean baseline to evaluate the controller's ask-or-answer decision boundary without inducing degenerate strategies (e.g., always asking or never asking). While real-world deployments may present skewed ambiguity distributions, a balanced split provides a rigorous benchmark for selective clarification.

\subsection{Mixed Dataset Construction Rules}
\dataset is constructed deterministically from an initial set of ambiguous VQA items. For each ambiguous item, we construct one additional non-ambiguous variant by modifying the question text:

\paragraph{Context Completion}
We add the missing relevant context explicitly to the original question, yielding a fully specified VQA instance. These examples are labeled as \texttt{no\_clarification\_needed}, since the answer can be determined directly from the image and provided context.

\paragraph{Irrelevant Context Injection}
We add additional contextual information that is syntactically plausible but does not resolve the underlying ambiguity. These examples preserve the original ambiguity and are labeled as \texttt{needs\_clarification}. Irrelevant context is selected to avoid introducing new ambiguities or revealing the correct answer.

This augmentation procedure results in a balanced mixture of ambiguous and non-ambiguous questions while preserving the original visual content.

\subsection{Examples}
\label{sec:dataset_examples}

\noindent Example 1 (Location Ambiguity)

\vspace{0.4em}

\begin{tabular}{@{}p{0.38\linewidth}p{0.66\linewidth}@{}}
\textit{Original question} &
\makecell[tl]{Is this behavior legal?} \\

\textit{Context-completed} &
\makecell[tl]{Is this behavior legal in\\\hspace*{0.0em}Germany?} \\

\textit{Irrelevant context} &
\makecell[tl]{Is this behavior legal\\\hspace*{0.0em}while wearing a blue\\\hspace*{0.0em} jacket?} \\
\end{tabular}

\vspace{1em}

\noindent Example 2 (Temporal Ambiguity)

\vspace{0.4em}

\begin{tabular}{@{}p{0.38\linewidth}p{0.66\linewidth}@{}}
\textit{Original question} &
\makecell[tl]{Can this vehicle be parked\\\hspace*{0.0em}here?} \\

\textit{Context-completed} &
\makecell[tl]{Can this vehicle be parked\\\hspace*{0.0em}here on Sundays?} \\

\textit{Irrelevant context} &
\makecell[tl]{Can this vehicle be parked\\\hspace*{0.0em}here during the afternoon?} \\
\end{tabular}

\section{Training Details}
\label{app:training_details}

\subsection{GRPO-CR Training}

We implemented \method on top of the open-sourced GRPO training framework from GRIT \citep{fan_2025_grit}. All GRPO experiments are conducted on 4$\times$A100 (80GB) GPUs using distributed training with DeepSpeed ZeRO-2 or ZeRO-3.

We train for a single epoch over the \dataset ambiguous set training split. Each training instance generates multiple clarification candidates, which are optimized using preference-based rewards described in Section~\ref{sec:clarification_learning}.
We restrict the interaction to a single clarification turn.

\paragraph{Optimization.}
We use the AdamW optimizer with a learning rate of $2\times10^{-6}$ and a cosine learning rate scheduler.
The per-device training batch size is 2 with gradient accumulation over 2 steps.
We set the KL regularization coefficient to $\beta=0.01$.
Training is performed in bfloat16 precision.

\paragraph{Generation and context limits.}
For each prompt, we sample 2 clarification generations.
The maximum prompt length is set to 1000 tokens and the maximum completion length to 768 tokens.

\paragraph{Evaluation and checkpointing.}
Evaluation is performed every 50 steps, and checkpoints are saved every 20 steps, keeping up to 5 checkpoints.
All training statistics and generated clarifications are logged to Weights \& Biases.

\subsection{Supervised Fine-Tuning}

For supervised baselines, we fine-tune the same backbone model using the MS-SWIFT framework \citep{zhao_2024_swifta}.
We adopt parameter-efficient fine-tuning with LoRA and train on the \dataset.

\paragraph{Setup.}
SFT is conducted on 2*A100(80GB) GPUs with DeepSpeed ZeRO-2.
We use a LoRA rank of 8 and $\alpha=32$, applying LoRA to all linear layers.
The model is trained for 3 epochs with a learning rate of $1\times10^{-4}$ and a warmup ratio of 0.05.

\paragraph{Batching and precision.}
The per-device batch size is 1 with gradient accumulation to achieve an effective batch size of 16.
All SFT experiments are trained in bfloat16 precision, with a maximum sequence length of 4096 tokens.

\paragraph{Evaluation and checkpointing.}
We evaluate and save checkpoints every 100 steps, keeping the two most recent checkpoints.

\subsection{In-Context Learning (ICL) Baseline Setup}
\label{app:icl_setup}
To evaluate whether improved prompting enhances zero-shot behavior, we construct a 4-shot ICL baseline. Each prompt augments the zero-shot instruction with:
\begin{itemize}[noitemsep, topsep=0pt, leftmargin=12pt]
    \item 1 context-sufficient example requiring a direct answer.
    \item 3 context-missing examples randomly sampled from the training set, covering \emph{Cultural Background}, \emph{Location \& Orientation}, and \emph{Temporal Information} ambiguity categories.
\end{itemize}
Each few-shot example provides the image, question, and ground-truth output (either direct answer or target clarification question). While ICL slightly improves over zero-shot prompting, it remains behind parameter-tuned models ($\text{CoA}_{\text{SFT}}$ and $\text{CoA}_{\text{SFT+GRPO}}$).

\begin{table*}[t!]
\centering
\small
\begin{tabular}{l l c}
\toprule
\textbf{Reward Component} & \textbf{Setting} & \textbf{Clarification Effectiveness} \\
\midrule

\multicolumn{3}{c}{\textbf{Ambiguity Reward Sweep}} \\
\midrule
Ambiguity & $+0.7, -0.5$ & $0.925 \pm 0.118$ \\
Ambiguity & $+0.5, -0.3$ (chosen) & $0.938 \pm 0.090$ \\
Ambiguity & $+0.3, -0.1$ & $0.938 \pm 0.039$ \\

\midrule
\multicolumn{3}{c}{\textbf{Novelty Reward Sweep}} \\
\midrule
Novelty & $+1.0, -0.1, -1.0$  & $0.730 \pm 0.093$ \\
Novelty & $+0.3, -0.1, -0.3$ (chosen) & $0.852 \pm 0.108$ \\
Novelty & $+0.1, +0.0, -0.1$ & $0.820 \pm 0.088$ \\

\midrule
\multicolumn{3}{c}{\textbf{Format Reward Sweep}} \\
\midrule
Format & $+1.0, -1.0$ & $0.710 \pm 0.205$ \\
Format & $+0.5, -0.5$ (chosen) & $0.750 \pm 0.090$ \\
Format & $+0.1, -0.1$ & $0.805 \pm 0.205$ \\

\midrule
\multicolumn{3}{c}{\textbf{Relevance Reward Sweep}} \\
\midrule
Relevance & $+0.5,\ +0.1,\ -0.3$ & $0.831 \pm 0.130$ \\
Relevance & $+0.3,\ +0.1,\ -0.1$ (chosen) & $0.831 \pm 0.030$ \\
Relevance & $+0.1,\ +0.0,\ -0.1$ & $0.796 \pm 0.090$ \\

\bottomrule
\end{tabular}
\caption{Parameter sweep over GRPO-CR reward coefficients. Each sweep varies one reward component while keeping all others fixed. Clarification effectiveness is computed using the ambiguity resolution reward and normalized to $[0,1]$ for comparability across settings. Results are reported on CtxClarify (Auto) using InternVL3-2B.}
\label{tab:reward_sweep}
\end{table*}

\section{Human Validation of Ambiguity Resolution}
\label{app:human_validation}

To validate that the proposed automatic ambiguity resolution reward reflects human judgment, we conduct a manual evaluation on a subset of the \dataset test set. Human annotations were performed by the authors.

\subsection{Sampling Protocol}
We randomly sample 30\% of the \dataset test set. For each sampled example, we evaluate clarification questions generated by three methods:vanilla prompting, supervised fine-tuning, and GRPO-CR. The same sampled examples are used across all methods and model variants to ensure a fair comparison.

\subsection{Annotation Task}
Human annotators are presented with the following information for each example:
\begin{itemize}[noitemsep, topsep=0pt]
    \item the image,
    \item the original ambiguous question,
    \item the model generated clarification question,
    \item the ground truth annotated clarification question, answer and final answer.
\end{itemize}

Annotators are asked to assess the extent to which the clarification question, when answered, resolves the ambiguity in the original question. Each clarification is assigned a graded score:
\begin{itemize}[noitemsep, topsep=0pt, leftmargin=12pt]
    \item \textbf{0 (Not Resolved):} The clarification does not address the source of ambiguity or would not change the final answer.
    \item \textbf{0.5 (Partially Resolved):} The clarification addresses part of the ambiguity but leaves multiple plausible answers.
    \item \textbf{1 (Fully Resolved):} The clarification directly targets the missing information and would uniquely determine the final answer.
\end{itemize}

\section{VQA Final Answer Accuracy Human Evaluation}
\label{app:human_vqa_eval}

We conduct a human evaluation to assess the correctness of the final VQA answers produced by different methods, focusing on whether the final response is reasonable for the given VQA. Human annotations were performed by the authors.

\subsection{Sampling Protocol}
We randomly sample 20\% of the test set predictions for human annotation. Sampling is performed uniformly across methods to ensure fair and unbiased comparison.

\subsection{Annotation Task}
For each evaluated example, annotators are provided with:
\begin{itemize}[noitemsep,topsep=0pt, leftmargin=18pt]
    \item the image,
    \item the original question,
    \item the model’s final answer,
    \item the clarification process, including the clarification question and response, when applicable
\end{itemize}

\noindent Each example is assigned a binary score:
\textbf{1} if the final answer is reasonable, and
\textbf{0} otherwise.

\subsection{LLM Judge Discrepancy Analysis}
\label{app:llm_judge_analysis}
Our human validation study indicates an 82\% agreement rate with the GPT-4o judge. Qualitative error inspection reveals that disagreements stem primarily from the LLM judge applying a stricter, more conservative evaluation standard than human annotators:
\begin{enumerate}[noitemsep, topsep=0pt, leftmargin=14pt]
    \item \textbf{Strict Formatting Constraints:} The LLM judge penalizes minor formatting variations. For example, answering ``1010'' for a 24-hour clock question was rejected by the LLM for missing colons, whereas human annotators accepted it as semantically equivalent to ``10:10''.
    \item \textbf{Overly Strict Evidence Requirements:} The LLM judge frequently penalizes responses relying on lightweight commonsense or conversational grounding. In a workplace query (``I am making a report to my leader''), the answer ``the person on the left is the leader'' was accepted by humans as reasonable contextual grounding, but rejected by the LLM for lacking explicit visual proof.
    \item \textbf{Conservative Domain Interpretation:} In financial queries where green text implied stock gains, human annotators accepted ``the stock is performing well'', while the LLM judge rejected it due to demanding additional external market context.
\end{enumerate}

\section{Additional Error Analysis}
\label{app:case_study}

Figure \ref{fig:case_studies_ambiguity} includes examples of ambiguous questions in our dataset, with the question and the ground-truth clarifying question, along with model responses for the prompting baseline and CoA trained with \method. In the vast majority of cases, the prompting baseline answers directly (incorrectly) without seeking additional clarification. The baseline also occasionally fails by reproducing a variant of the original question. For CoA, the model asks a relevant clarification question, which resolves the ambiguity in some cases. Figure \ref{fig:case_studies_clear} shows case studies for non-ambiguous questions from our contrast set.

\section{Prompt Templates}
\label{app:prompt_templates}

We reproduce the prompts used for ambiguity detection, clarification question generation, and the end-to-end prompt baseline below. 

\subsection{Ambiguity Detection}
\small
\begin{verbatim}
System:
You are a visual Q&A ambiguity checker. Given one image and one question 
about that image, decide if the question needs further clarification 
before it can be reliably answered from pixels. Mark `yes` if answering 
depends on off-image knowledge. Otherwise, mark `no`.  

Output only a single token: `yes` or `no`.

User: 
<Image>
Question: {question}

\end{verbatim}
\normalsize

\subsection{Clarification Question Generation}

\small
\begin{verbatim}
System:
You write exactly one clarifying question. Do not answer the original 
question.

Given an image and an original question, ask for the single missing fact 
that would determine the answer.

Input: (1) image, (2) original question.
Output: exactly one short clarification question without any prefix.

User:
<Image>
Question: {question}
\end{verbatim}
\normalsize

\subsection{Final Answer Generation}
\small
\begin{verbatim}
System:
You are a visual assistant. Answer the user's question using the image.
Input: (1) image, (2) original question.
Output: final answer

User:
<Image>
Question: {question}
\end{verbatim}
\normalsize

\subsection{Vanilla Prompt}

\small
\begin{verbatim}
System:
You are a visual assistant. You answer the question directly whenever 
the image provides enough information. Only ask a clarification question 
if the answer would differ depending on missing information.

Input: (1) image, (2) original question.
Output: final answer or a single short clarification question

User:
<Image>
Question: {question}
\end{verbatim}
\normalsize

\subsection{VQA Accuracy LLM Judge}

\small
\begin{verbatim}
System:
You are a strict but fair VQA (Visual Question Answering) judge.

Your job: decide whether the provided answer(s) correctly answer the 
given question and image.

Rules:
- Output ONLY valid JSON, no preamble, no markdown fences.
- Output format: {"verdict": "correct" | "incorrect", 
"reasoning": "<one sentence>"}
- "correct"   → the answer is a valid, reasonable response to the 
question and image pair.
- "incorrect" → the answer is wrong, irrelevant, or does not address the
question and image pair.
- Be strict about factual accuracy; do not give credit for vague 
non-answers.

User:
<Image>
Question: {question}
Answer: {answer}
\end{verbatim}
\normalsize

\end{document}

%% file: commands.tex
\newcommand\dataset{\textsc{ContextClarify}\xspace}

\newcommand\method{GRPO-CR\xspace}

\newcolumntype{L}[1]{>{\raggedright\let\newline\\\arraybackslash\hspace{0pt}}p{#1}}
\newcolumntype{R}[1]{>{\raggedleft\let\newline\\\arraybackslash\hspace{0pt}}p{#1}}
\newcolumntype{M}[1]{>{\raggedright\let\newline\\\arraybackslash\hspace{0pt}}m{#1}}

\definecolor{darkgreen}{rgb}{0.0, 0.4, 0.13}
\definecolor{darkgreen2}{HTML}{037171}
\definecolor{lightgreen2}{HTML}{83c5be}
\definecolor{coral2}{HTML}{e29578}
\definecolor{lightlightblue}{rgb}{0.9, 0.95, 1.0}

\definecolor{mustard}{rgb}{0.9, .61, .11}

%% file: tables/ctxclarify_results_table.tex
\begin{table}[t]
    \vspace{-3mm}
    \centering
    \scriptsize
    \setlength{\tabcolsep}{2pt}
    \renewcommand{\arraystretch}{0.90}

    \begin{tabular}{lcccccc}
        \toprule
        \textbf{Model} & \textbf{Attributes} & \textbf{Cultural} & \textbf{Location \&} & \textbf{Relation-} & \textbf{Temporal} & \textbf{Overall} \\
         & & \textbf{Backgr.} & \textbf{Orient.} & \textbf{ships} & \textbf{Info.} & \\
        \midrule
        \rowcolor[gray]{.95} \multicolumn{7}{l}{\textit{Closed-weight models (zero-shot)}} \\
        \midrule
        GPT-4o & 0.611$\pm$0.138 & 0.690$\pm$0.102 & 0.630$\pm$0.080 & 0.714$\pm$0.177 & 0.483$\pm$0.072 & 0.619$\pm$0.030 \\        
        Gemini-2.5-flash & 0.630$\pm$0.080 & 0.571$\pm$0.177 & 0.556$\pm$0.138 & 0.690$\pm$0.102 & 0.383$\pm$0.072 & 0.560$\pm$0.030 \\        \midrule
        \rowcolor[gray]{.95} \multicolumn{7}{l}{\textit{Open-weight models (baselines and finetuned on \dataset using CoA framework)}} \\
        \midrule
        InternVL3-2B (Zero-shot) & 0.093$\pm$0.080 & 0.048$\pm$0.102 & 0.056$\pm$0.138 & 0.143$\pm$0.177 & 0.033$\pm$0.072 & 0.087$\pm$0.017 \\
        InternVL3-2B (AA$_{\text{GRPO}}$) & 0.278$\pm$0.138 & 0.119$\pm$0.102 & \textbf{0.407$\pm$0.080} & 0.143$\pm$0.177 & 0.217$\pm$0.072 & 0.238$\pm$0.030 \\
        InternVL3-2B (ICL) & 0.130$\pm$0.080 & 0.143$\pm$0.177 & 0.167$\pm$0.138 & 0.095$\pm$0.102 & 0.100$\pm$0.124 & 0.123$\pm$0.017 \\
        InternVL3-2B (CoA$_{\text{SFT}}$) & 0.185$\pm$0.080 & 0.119$\pm$0.102 & 0.130$\pm$0.080 & 0.190$\pm$0.102 & 0.083$\pm$0.072 & 0.127$\pm$0.017 \\
        InternVL3-2B (CoA$_{\text{SFT+GRPO}}$) & \textbf{0.389$\pm$0.138} & \textbf{0.310$\pm$0.102} & 0.204$\pm$0.080 & \textbf{0.286$\pm$0.177} & \textbf{0.367$\pm$0.072} & \textbf{0.313$\pm$0.017} \\
        \noalign{\vskip 1pt}
        \hdashline[0.5pt/2pt]
        \noalign{\vskip 2pt}
        InternVL3-2B (CoA$_{\text{Oracle}}$) & 0.389$\pm$0.138 & 0.333$\pm$0.102 & 0.296$\pm$0.080 & 0.357$\pm$0.177 & 0.400$\pm$0.124 & 0.361$\pm$0.017 \\        \midrule
        Qwen2.5-3B (Zero-shot) & 0.315$\pm$0.080 & 0.095$\pm$0.102 & 0.278$\pm$0.138 & \textbf{0.381$\pm$0.102} & 0.100$\pm$0.124 & 0.242$\pm$0.017 \\
        Qwen2.5-3B (AA$_{\text{GRPO}}$) & 0.278$\pm$0.138 & 0.286$\pm$0.177 & 0.296$\pm$0.080 & 0.262$\pm$0.102 & \textbf{0.350$\pm$0.124} & 0.321$\pm$0.030 \\        
        Qwen2.5-3B (ICL) & 0.352$\pm$0.080 & 0.143$\pm$0.177 & 0.278$\pm$0.138 & \textbf{0.381$\pm$0.102} & 0.233$\pm$0.072 & 0.274$\pm$0.030 \\        
        Qwen2.5-3B (CoA$_{\text{SFT}}$) & \textbf{0.426$\pm$0.080} & 0.310$\pm$0.102 & 0.352$\pm$0.080 & 0.143$\pm$0.177 & \textbf{0.350$\pm$0.124} & 0.321$\pm$0.030 \\
        Qwen2.5-3B (CoA$_{\text{SFT+GRPO}}$) & 0.389$\pm$0.138 & \textbf{0.381$\pm$0.102} & \textbf{0.463$\pm$0.080} & 0.310$\pm$0.102 & \textbf{0.350$\pm$0.124} & \textbf{0.381$\pm$0.030} \\
        \noalign{\vskip 1pt}
        \hdashline[0.5pt/2pt]
        \noalign{\vskip 2pt}
        Qwen2.5-3B (CoA$_{\text{Oracle}}$) & 0.444$\pm$0.138 & 0.405$\pm$0.102 & 0.500$\pm$0.138 & 0.310$\pm$0.102 & 0.400$\pm$0.124 & 0.421$\pm$0.017 \\        
        \midrule
        Qwen2.5-7B (Zero-shot) & 0.259$\pm$0.080 & 0.452$\pm$0.102 & 0.315$\pm$0.080 & 0.262$\pm$0.102 & 0.083$\pm$0.072 & 0.266$\pm$0.017 \\
        Qwen2.5-7B (AA$_{\text{GRPO}}$) & 0.333$\pm$0.138 & 0.405$\pm$0.102 & \textbf{0.389$\pm$0.138} & 0.286$\pm$0.177 & 0.367$\pm$0.072 & 0.357$\pm$0.030 \\        
        Qwen2.5-7B (ICL) & 0.148$\pm$0.080 & 0.357$\pm$0.177 & 0.278$\pm$0.138 & 0.286$\pm$0.177 & 0.083$\pm$0.072 & 0.230$\pm$0.017 \\        
        Qwen2.5-7B (CoA$_{\text{SFT}}$) & 0.296$\pm$0.080 & 0.524$\pm$0.102 & 0.352$\pm$0.080 & 0.310$\pm$0.102 & 0.250$\pm$0.124 & 0.317$\pm$0.017 \\
        Qwen2.5-7B (CoA$_{\text{SFT+GRPO}}$) & \textbf{0.389$\pm$0.138} & \textbf{0.571$\pm$0.177} & \textbf{0.389$\pm$0.138} & \textbf{0.333$\pm$0.102} & \textbf{0.400$\pm$0.124} & \textbf{0.409$\pm$0.017} \\
        \noalign{\vskip 1pt}
        \hdashline[0.5pt/2pt]
        \noalign{\vskip 2pt}
        Qwen2.5-7B (CoA$_{\text{Oracle}}$) & 0.444$\pm$0.138 & 0.571$\pm$0.177 & 0.444$\pm$0.138 & 0.429$\pm$0.177 & 0.417$\pm$0.072 & 0.452$\pm$0.030 \\        
        \bottomrule
    \end{tabular}
    \caption{VQA accuracy on \dataset. Our CoA-tuned models improve over their base counterparts across all question categories, in some cases approaching  performance of frontier VLMs.}
    \label{tab:model_results}
\end{table}

%% file: tables/ood_results_table.tex
\begin{wraptable}{r}{0.4\linewidth}
\centering
\scriptsize
\setlength{\tabcolsep}{2pt}
\vspace{-2mm}
\begin{tabular}{l ccc}
\toprule
\textbf{Method} & \textbf{IVL-2B} & \textbf{Qw-3B} & \textbf{Qw-7B} \\
\midrule
\textit{ClearVQA} & & & \\
Zero-shot & 0.015 & 0.075 & 0.083 \\
AA$_{\text{GRPO}}$ & 0.150 & 0.175 & 0.205 \\
ICL & 0.085 & 0.098 & 0.135 \\
CoA$_{SFT}$ & 0.138 & 0.143 & 0.165 \\
CoA$_{SFT+GRPO}$ & \textbf{0.213} & \textbf{0.253} & \textbf{0.345} \\
\noalign{\vskip 1pt}
\hdashline[0.5pt/2pt]
\noalign{\vskip 2pt}
CoA$_{\text{Oracle}}$ & 0.305 & 0.358 & 0.418 \\
\midrule
\textit{VizWiz} & & & \\
Zero-shot & 0.040 & 0.165 & 0.225 \\
AA$_{\text{GRPO}}$ & 0.310 & \textbf{0.380} & \textbf{0.395} \\
ICL & 0.090 & 0.185 & 0.215 \\
CoA$_{SFT}$ & 0.190 & 0.310 & 0.305 \\
CoA$_{SFT+GRPO}$ & \textbf{0.315} & 0.360 & 0.375 \\
\noalign{\vskip 1pt}
\hdashline[0.5pt/2pt]
\noalign{\vskip 2pt}
CoA$_{\text{Oracle}}$ & 0.310 & 0.380 & 0.395 \\
\bottomrule
\end{tabular}
\caption{OOD perf on ClearVQA and VizWiz. VizWiz contains only ambiguous VQAs, so AA (Always Ask) is equivalent to Oracle.}
\label{tab:vqa_ood_results}
\vspace{-3mm}
\end{wraptable}

%% file: figures/multiturn_vqa_accuracy.tex
\begin{figure*}[t]
\vspace{-4mm}
\centering
\resizebox{0.98\linewidth}{!}{
    \begin{subfigure}[b]{0.33\linewidth}
        \centering
        \begin{tikzpicture}
        \begin{axis}[
            width=\linewidth,
            height=4cm,
            xlabel={Turns},
            ylabel={VQA Accuracy},
            xmin=0, xmax=4,
            xtick={0,1,2,3,4},
            ymin=0, ymax=0.45,
            grid=both,
            line width=1pt,
            title={\dataset Dataset}
        ]
        \addplot+[color=black, mark=o] coordinates {(1,0.2619) (2,0.2850) (3,0.2930)};
        \addplot+[color=blue, mark=square] coordinates {(1,0.3214) (2,0.3400) (3,0.3425)};
        \addplot+[color=red, mark=triangle] coordinates {(1,0.4167) (2,0.4250) (3,0.4250)};
        \end{axis}
        \end{tikzpicture}
    \end{subfigure}
    \begin{subfigure}[b]{0.33\linewidth}
        \centering
        \begin{tikzpicture}
        \begin{axis}[
            width=\linewidth,
            height=4cm,
            xlabel={Turns},
            ylabel={}, 
            xmin=0, xmax=4,
            xtick={0,1,2,3,4},
            ymin=0, ymax=0.45,
            grid=both,
            line width=1pt,
            title={ClearVQA Dataset}
        ]
        \addplot+[color=black, mark=o] coordinates {(1,0.0825) (2,0.1250) (3,0.1250)};
        \addplot+[color=blue, mark=square] coordinates {(1,0.1650) (2,0.1925) (3,0.2050)};
        \addplot+[color=red, mark=triangle] coordinates {(1,0.3450) (2,0.3595) (3,0.3700)};
        \end{axis}
        \end{tikzpicture}
    \end{subfigure}
    \begin{subfigure}[b]{0.33\linewidth}
        \centering
        \begin{tikzpicture}
        \begin{axis}[
            width=\linewidth,
            height=4cm,
            xlabel={Turns},
            ylabel={},
            xmin=0, xmax=4,
            xtick={0,1,2,3,4},
            ymin=0, ymax=0.45,
            grid=both,
            line width=1pt,
            title={VizWiz Dataset}
        ]
        \addplot+[color=black, mark=o] coordinates {(1,0.2250) (2,0.2450) (3,0.2550)};
        \addplot+[color=blue, mark=square] coordinates {(1,0.3050) (2,0.3200) (3,0.3050)};
        \addplot+[color=red, mark=triangle] coordinates {(1,0.3750) (2,0.3800) (3,0.3750)};
        \end{axis}
        \end{tikzpicture}
    \end{subfigure}
} 

    \vspace{-1mm}
    \begin{tikzpicture}
        \begin{axis}[
            hide axis,
            xmin=0, xmax=1, ymin=0, ymax=1,
            legend style={draw=none, legend columns=3, font=\small},
        ]
        \addlegendimage{color=black, mark=o}
        \addlegendentry{Prompt}
        \addlegendimage{color=blue, mark=square}
        \addlegendentry{CoA$_{SFT}$}
        \addlegendimage{color=red, mark=triangle}
        \addlegendentry{CoA$_{SFT+GRPO}$}
        \end{axis}
    \end{tikzpicture}
\vspace{-5.0cm}
\caption{VQA accuracy comparison shows the most significant gains occurring within the first turn.} 
\vspace{-2mm}
\label{fig:multi_turn_vqa}
\end{figure*}

%% file: figures/case_studies_main.tex
\begin{figure*}[t!]
\centering
\renewcommand{\arraystretch}{1.10}
\setlength{\tabcolsep}{3pt}
\vspace{-2mm}
\begin{minipage}[t]{0.48\textwidth}
\vspace{0pt}
\scriptsize
\colorbox{gray!8}{%
\parbox[t][5.0cm][t]{\linewidth}{%
\vspace{3pt}
\textbf{Example A} \\ [0.5mm]
\textbf{Q:} My mother is walking with her families. Which one is my mother? \\ [0.5mm]
\textbf{GTCQ:} Can you tell me who the other person is in relation to your mother?

\noindent
\begin{minipage}{0.28\linewidth}
  \centering
  \vfill
  \includegraphics[width=\linewidth]{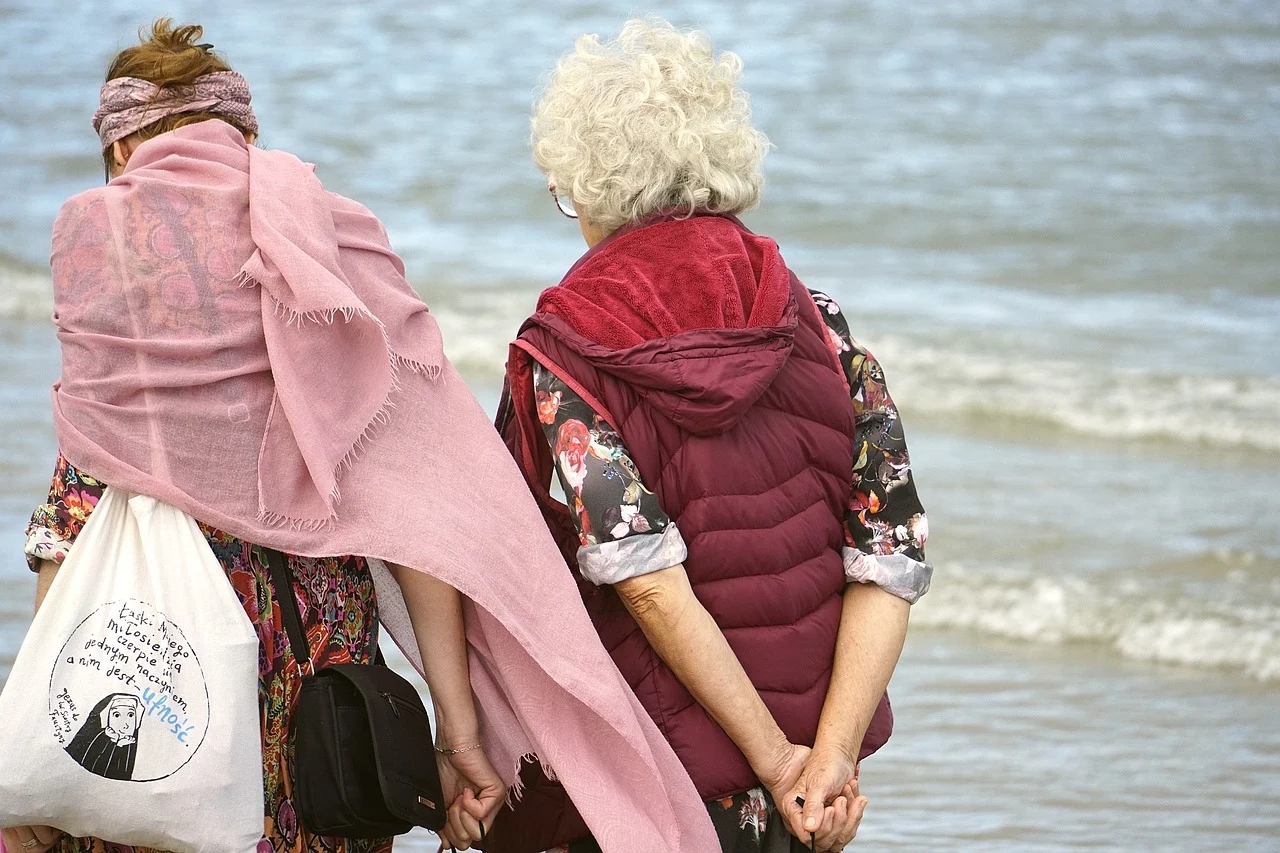}
  \vfill
\end{minipage}\hfill
\begin{minipage}[t]{0.70\linewidth}
  \scriptsize
\setlength{\tabcolsep}{2pt}
\renewcommand{\arraystretch}{1.15}

\begin{tabular}{@{}p{0.10\linewidth} p{0.75\linewidth}@{}}
\textbf{BL} &
\begin{tabular}{@{}l p{0.90\linewidth}@{}}
\textbf{A:}  & \textcolor{red}{The woman on the right is holding hands with the other woman, so she is likely the mother.} \textit{(unwarranted inference)}\\
\end{tabular}
\\[-4pt]
& \rule{\linewidth}{0.2pt} \\[-2pt]

\textbf{CoA} &
\begin{tabular}{@{}l p{0.90\linewidth}@{}}
\textbf{CQ:} & \textcolor{green!50!black}{Which family member is walking with your mother?} \textit{(good: resolves relationship)}\\
\textbf{CR:} & My grandmother. \\
\textbf{A:}  & \textcolor{green!50!black}{The left one.} \textit{(good: reasonable answer under given context)}\\
\end{tabular}
\\[-4pt]
\end{tabular}

\end{minipage}
}%
}%
\end{minipage}
\hfill
\begin{minipage}[t]{0.48\textwidth}
\vspace{0pt}
\scriptsize
\colorbox{gray!8}{%
\parbox[t][5.0cm][t]{\linewidth}{%
\vspace{3pt}
\textbf{Example B} \\ [0.5mm]
\textbf{Q:} I am driving on the left lane. Is it allowed? \\ [0.5mm]
\textbf{GTCQ:} That depends on the country's traffic laws. In which country was this photo taken? 

\noindent
\begin{minipage}{0.28\linewidth}
  \centering
  \vfill
  \includegraphics[width=\linewidth]{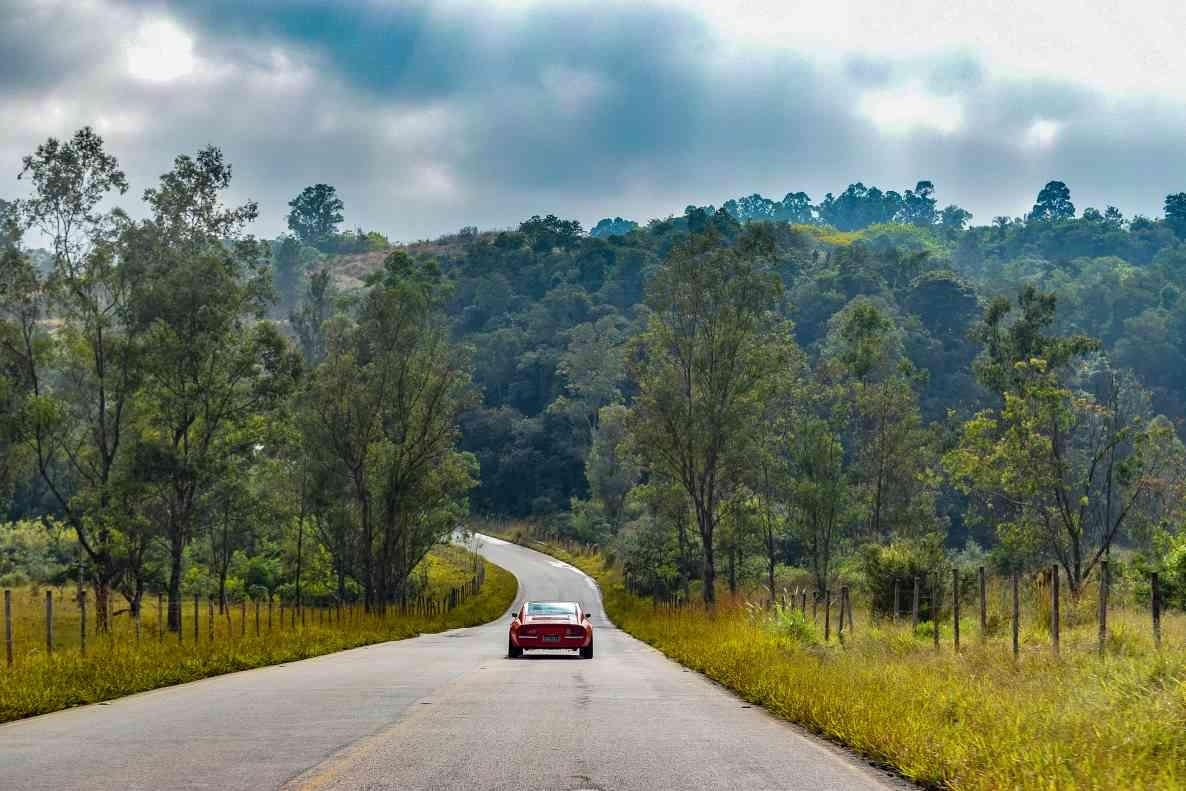}
  \vfill
\end{minipage}\hfill
\begin{minipage}[t]{0.70\linewidth}
\scriptsize
\setlength{\tabcolsep}{2pt}
\renewcommand{\arraystretch}{1.15}

\begin{tabular}{@{}p{0.10\linewidth} p{0.75\linewidth}@{}}

\textbf{BL} &
\begin{tabular}{@{}l p{0.90\linewidth}@{}}
\\[0.5pt]
\textbf{A:}  & \textcolor{red}{Yes, you are driving in the left lane allowed.} \textit{(unwarranted inference)} \\
\end{tabular}
\\[-4pt]
& \rule{\linewidth}{0.2pt} \\[-2pt]

\textbf{CoA} &
\begin{tabular}{@{}l p{0.90\linewidth}@{}}
\textbf{CQ:} & \textcolor{orange!80!black}{Is the road in the image a two-way or one-way road?} \textit{(partially relevant but does
not ask about the key disambiguator)} \\
\textbf{CR:} & Two-way road. \\
\textbf{A:}  & \textcolor{orange!80!black}{No it is not allowed.} \textit{(unwarranted inference based on partial context)}\\ [2pt]
\end{tabular}
\\
\end{tabular}
\end{minipage}
}%
}%
\end{minipage}
\caption{Select examples comparing responses from the prompting baseline (BL) and CoA$_{SFT+GRPO}$. In both examples, the BL answers without resolving the ambiguity. In Example A, CoA correctly resolves the ambiguity; in example B, while CoA produces a partially relevant clarifying question, it does not arrive at the correct answer. Abbreviations: Q = original question; GTCQ = ground truth clarification question; CQ = clarification question; CR = clarification response; A = final answer.}
\label{fig:case_studies_main_paper}
\end{figure*}

%% file: tables/module_level_results.tex
\begin{table}[t]
\centering
\begin{minipage}[t]{0.46\textwidth}
\centering
\scriptsize
\setlength{\tabcolsep}{1pt}
\renewcommand{\arraystretch}{0.88}
\begin{tabular}{l l l c c c c}
\toprule
Data & Model & Method & Acc $\uparrow$ & Prec & Recall $\uparrow$ & F1 $\uparrow$ \\
\midrule
\multirow{6}{*}{CtxClarify}
 & \multirow{2}{*}{IVL-2B}
   & Prompt & 0.500 & 0.000 & 0.000 & 0.000 \\
 & & SFT    & \textbf{0.536} & \textbf{0.588} & \textbf{0.238} & \textbf{0.339} \\
\cmidrule(lr){2-7}
 & \multirow{2}{*}{Qw-3B}
   & Prompt & 0.476 & 0.474 & 0.429 & 0.450 \\
 & & SFT    & \textbf{0.500} & \textbf{0.500} & \textbf{0.500} & \textbf{0.500} \\
\cmidrule(lr){2-7}
 & \multirow{2}{*}{Qw-7B}
   & Prompt & 0.595 & \textbf{0.667} & 0.381 & 0.485 \\
 & & SFT    & \textbf{0.702} & 0.660 & \textbf{0.833} & \textbf{0.737} \\
\midrule
\multirow{6}{*}{\makecell{ClearVQA\\(OOD)}}
 & \multirow{2}{*}{IVL-2B}
   & Prompt & \textbf{0.495} & 0.450 & 0.045 & 0.082 \\
 & & SFT    & 0.490 & \textbf{0.488} & \textbf{0.395} & \textbf{0.437} \\
\cmidrule(lr){2-7}
 & \multirow{2}{*}{Qw-3B}
   & Prompt & 0.423 & 0.432 & 0.500 & 0.464 \\
 & & SFT    & \textbf{0.470} & \textbf{0.476} & \textbf{0.605} & \textbf{0.533} \\
\cmidrule(lr){2-7}
 & \multirow{2}{*}{Qw-7B}
   & Prompt & 0.474 & 0.482 & 0.729 & 0.580 \\
 & & SFT    & \textbf{0.530} & \textbf{0.519} & \textbf{0.805} & \textbf{0.631} \\
\bottomrule
\end{tabular}
\caption{\textbf{Controller performance} on \dataset and ClearVQA. Prompt = zero-shot prompting. Training the CoA controller with SFT improves ambiguity detection on \dataset over Prompt and performance gains generalize to ClearVQA.}
\label{tab:controller_perf}
\end{minipage}
\hfill
\begin{minipage}[t]{0.50\textwidth}
\centering
\small
\setlength{\tabcolsep}{1pt}
\begin{tabular}{l l c c c}
\toprule
Dataset & Method & IVL-2B & Qw-3B & Qw-7B \\
\midrule
\multirow{3}{*}{\makecell{CtxClarify}}
& Prompt & 0.476 & 0.518 & 0.571 \\
& CoA$_{SFT}$    & 0.593 & 0.643 & 0.643 \\
& CoA$_{SFT+GRPO}$ & \textbf{0.685} & \textbf{0.768} & \textbf{0.810} \\
\midrule
\multirow{3}{*}{\makecell{ClearVQA \\(OOD)}}
& Prompt & 0.526 & 0.561 & 0.660 \\
& CoA$_{SFT}$    & 0.648 & 0.670 & 0.738 \\
& CoA$_{SFT+GRPO}$ & \textbf{0.653} & \textbf{0.796} & \textbf{0.804} \\
\midrule
\multirow{3}{*}{\makecell{VizWiz \\(OOD)}}
& Prompt & 0.504 & 0.453 & 0.689 \\
& CoA$_{SFT}$   & 0.611 & 0.559 & 0.708 \\
& CoA$_{SFT+GRPO}$ & \textbf{0.764} & \textbf{0.720} & \textbf{0.864} \\
\bottomrule
\end{tabular}
\caption{\textbf{Clarification question quality} across datasets and models.
Scores range from $0$ to $1$. While SFT baseline provides a notable boost in precision over zero-shot prompting, our \method method consistently achieves the highest clarification scores across both in and out domain datasets.}
\label{tab:clarification_policy_combined}
\end{minipage}
\vspace{-4pt}
\end{table}

%% file: tables/human_eval_vqa.tex
\begin{wrapfigure}{r}{0.4\textwidth}
\centering
\scriptsize
\setlength{\tabcolsep}{4pt}
\vspace{-6mm}

\begin{tabular}{l ccc}
\toprule
\textbf{Method} & \textbf{IVL-2B} & \textbf{Qw-3B} & \textbf{Qw-7B} \\
\midrule
\multicolumn{4}{l}{\textit{CtxClarify}} \\
Zero-shot
    & 0.105
    & 0.236
    & 0.316 \\
CoA$_{SFT+GRPO}$
    & \textbf{0.368}
    & \textbf{0.316}
    & \textbf{0.474} \\
\midrule
\multicolumn{4}{l}{\textit{ClearVQA}} \\
Zero-shot
    & 0.276
    & 0.233
    & 0.263 \\
CoA$_{SFT+GRPO}$
    & \textbf{0.333}
    & \textbf{0.341}
    & \textbf{0.375} \\
\bottomrule
\end{tabular}

\captionof{table}{Human evaluation of VQA accuracy on CtxClarify and ClearVQA.}
\label{tab:human_eval_vqa}

\vspace{3mm}

\begin{tabular}{l ccc}
\toprule
\textbf{Method} & \textbf{IVL-2B} & \textbf{Qw-3B} & \textbf{Qw-7B} \\
\midrule
Zero-shot
    & 0.104
    & 0.376
    & 0.500 \\
CoA$_{SFT}$
    & 0.562
    & 0.712
    & 0.712 \\
CoA$_{SFT+GRPO}$
    & \textbf{0.688}
    & \textbf{0.838}
    & \textbf{0.876} \\
\bottomrule
\end{tabular}

\captionof{table}{Human evaluation of clarification question quality.
Questions generated by CoA$_{SFT+GRPO}$ on CtxClarify consistently achieve higher human evaluation scores
across models.}
\label{tab:clarification_policy_human}
\end{wrapfigure}



%% file: tables/appendix_case_studies.tex
\begin{figure*}[t!]
\centering
\renewcommand{\arraystretch}{1.10}
\setlength{\tabcolsep}{3pt}

\begin{minipage}[t]{0.48\textwidth}
\vspace{0pt}
\scriptsize
\colorbox{gray!8}{%
\parbox[t][4.8cm][t]{\linewidth}{%
\vspace{3pt}
\textbf{(a) Q:} My mother is walking with her families. Which one is my mother? \\
\textbf{GTCQ:} Can you tell me who the other person is in relation to your mother?

\vspace{3pt}
\noindent
\begin{minipage}{0.28\linewidth}
  \centering
  \vfill
  \includegraphics[width=\linewidth]{figs/mother.jpg}
  \vfill
\end{minipage}\hfill
\begin{minipage}[t]{0.70\linewidth}
  \scriptsize
\setlength{\tabcolsep}{2pt}
\renewcommand{\arraystretch}{1.15}

\begin{tabular}{@{}p{0.10\linewidth} p{0.75\linewidth}@{}}

\textbf{BL} &
\begin{tabular}{@{}l p{0.90\linewidth}@{}}
\textbf{A:}  & \textcolor{red}{The woman on the right is holding hands with the other woman, so she is likely the mother.} \textit{(unwarranted inference)}\\
\end{tabular}
\\[-4pt]
& \rule{\linewidth}{0.2pt} \\[-2pt]

\textbf{CoA} &
\begin{tabular}{@{}l p{0.90\linewidth}@{}}
\textbf{CQ:} & \textcolor{green!50!black}{Which family member is walking with your mother?} \textit{(good: resolves relationship)}\\
\textbf{CR:} & My grandmother. \\
\textbf{A:}  & \textcolor{green!50!black}{The left one.} \textit{(good: reasonable answer under given context)}\\
\end{tabular}
\\[-4pt]
\end{tabular}

\end{minipage}
}%
}%
\end{minipage}
\hfill
\begin{minipage}[t]{0.48\textwidth}
\vspace{0pt}
\scriptsize
\colorbox{gray!8}{%
\parbox[t][4.8cm][t]{\linewidth}{%
\vspace{3pt}
\textbf{(b) Q:} I am driving on the left lane. Is it allowed? \\
\textbf{GTCQ:} That depends on the country's traffic laws. In which country was this photo taken? 

\vspace{8pt}
\noindent
\begin{minipage}{0.28\linewidth}
  \centering
  \vfill
  \includegraphics[width=\linewidth]{figs/left_lane.jpg}
  \vfill
\end{minipage}\hfill
\begin{minipage}[t]{0.70\linewidth}
\scriptsize
\setlength{\tabcolsep}{2pt}
\renewcommand{\arraystretch}{1.15}

\begin{tabular}{@{}p{0.10\linewidth} p{0.75\linewidth}@{}}

\textbf{BL} &
\begin{tabular}{@{}l p{0.90\linewidth}@{}}
\textbf{A:}  & \textcolor{red}{Yes, you are driving in the left lane allowed.} \textit{(unwarranted inference)} \\
\end{tabular}
\\[-2pt]
& \rule{\linewidth}{0.2pt} \\[-2pt]

\textbf{CoA} &
\begin{tabular}{@{}l p{0.90\linewidth}@{}}
\textbf{CQ:} & \textcolor{orange!80!black}{Is the road in the image a two-way or one-way road?} \textit{(partially relevant but does
not ask about the key disambiguator)} \\
\textbf{CR:} & Two-way road. \\
\textbf{A:}  & \textcolor{orange!80!black}{No it is not allowed.} \textit{(unwarranted inference based on partial context)}\\
\end{tabular}
\\[-4pt]
\end{tabular}
\end{minipage}
}%
}%
\end{minipage}

\vspace{4pt}

\begin{minipage}[t]{0.48\textwidth}
\vspace{0pt}
\scriptsize
\colorbox{gray!8}{%
\parbox[t][4.0cm][t]{\linewidth}{%
\vspace{3pt}
\textbf{(c) Q:} Am I going from left to right or from right to left? \\
\textbf{GTCQ:} Is the door opening towards you or away from you?

\vspace{8pt}
\noindent
\begin{minipage}{0.28\linewidth}
  \centering
  \vfill
  \includegraphics[width=\linewidth]{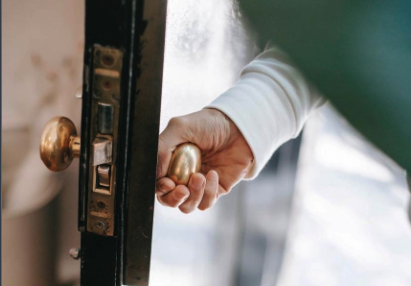}
  \vfill
\end{minipage}\hfill
\begin{minipage}[t]{0.70\linewidth}
\scriptsize
\setlength{\tabcolsep}{2pt}
\renewcommand{\arraystretch}{1.15}

\begin{tabular}{@{}p{0.10\linewidth} p{0.75\linewidth}@{}}

\textbf{BL} &
\begin{tabular}{@{}l p{0.90\linewidth}@{}}
\textbf{A:}  & \textcolor{red}{You are going from left to right.} \textit{(unwarranted inference)}\\
\end{tabular}
\\[-4pt]
& \rule{\linewidth}{0.2pt} \\[-2pt]

\textbf{CoA} &
\begin{tabular}{@{}l p{0.90\linewidth}@{}}
\textbf{CQ:} & \textcolor{green!50!black}{Is the door opening towards you or away from you?} \textit{(good: resolves temporal ambiguity)} \\
\textbf{CR:} & Opening towards me. \\
\textbf{A:}  & \textcolor{green!50!black}{From left to right}. \textit{(good: reasonable answer under given context)} \\
\end{tabular}
\\[-4pt]

\end{tabular}
\end{minipage}
}%
}%
\end{minipage}
\hfill
\begin{minipage}[t]{0.48\textwidth}
\vspace{0pt}
\scriptsize
\colorbox{gray!8}{%
\parbox[t][4.0cm][t]{\linewidth}{%
\vspace{3pt}
\textbf{(d) Q:} I want to buy two bottles of milk. Is my gesture appropriate? \\
\textbf{GTCQ:} In what country or region is this gesture being used?

\vspace{3pt}
\noindent
\begin{minipage}{0.28\linewidth}
  \centering
  \vfill
  \includegraphics[width=\linewidth]{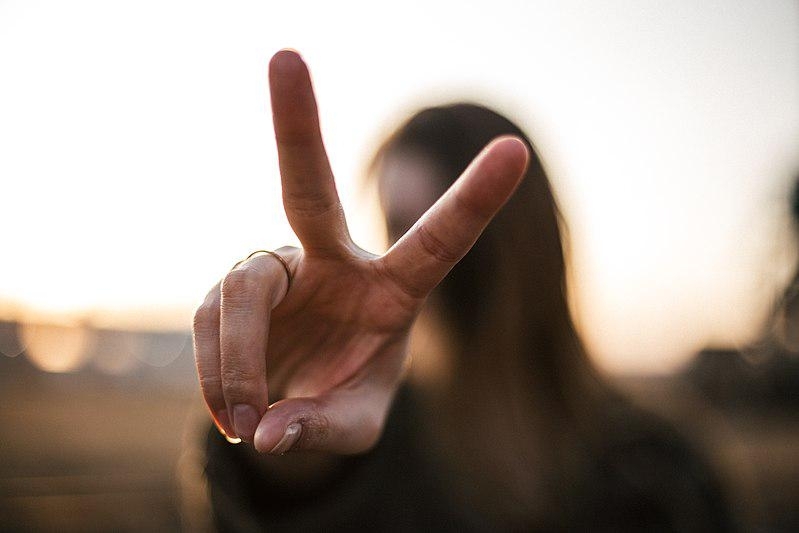}
  \vfill
\end{minipage}\hfill
\begin{minipage}[t]{0.70\linewidth}
\scriptsize
\setlength{\tabcolsep}{2pt}
\renewcommand{\arraystretch}{1.15}

\begin{tabular}{@{}p{0.10\linewidth} p{0.75\linewidth}@{}}

\textbf{BL} &
\begin{tabular}{@{}l p{0.90\linewidth}@{}}
\textbf{A:}  & \textcolor{red}{Yes, your gesture is appropriate} \textit{(unwarranted inference)} \\
\end{tabular}
\\[-2pt]
& \rule{\linewidth}{0.2pt} \\[-2pt]

\textbf{CoA} &
\begin{tabular}{@{}l p{0.90\linewidth}@{}}
\textbf{CQ:} & \textcolor{green!50!black}{What is the context or location where the gesture is being made?} \textit{(resolves the culture ambiguity)}\\
\textbf{CR:} & In Greece. \\
\textbf{A:}  & \textcolor{green!50!black}{No.} \textit{(good: resonable answer under the given context)} \\
\end{tabular}
\\[-2pt]

\end{tabular}

\end{minipage}
}%
}%
\end{minipage}

\vspace{4pt}

\begin{minipage}[t]{0.48\textwidth}
\vspace{0pt}
\scriptsize
\colorbox{gray!8}{%
\parbox[t][5.2cm][t]{\linewidth}{%
\vspace{3pt}
\textbf{(e) Q:} Is the subway going in the same direction as I face? \\
\textbf{GTCQ:} Is the subway train moving towards or away from you?

\vspace{3pt}
\noindent
\begin{minipage}{0.28\linewidth}
  \centering
  \vfill
  \includegraphics[width=\linewidth]{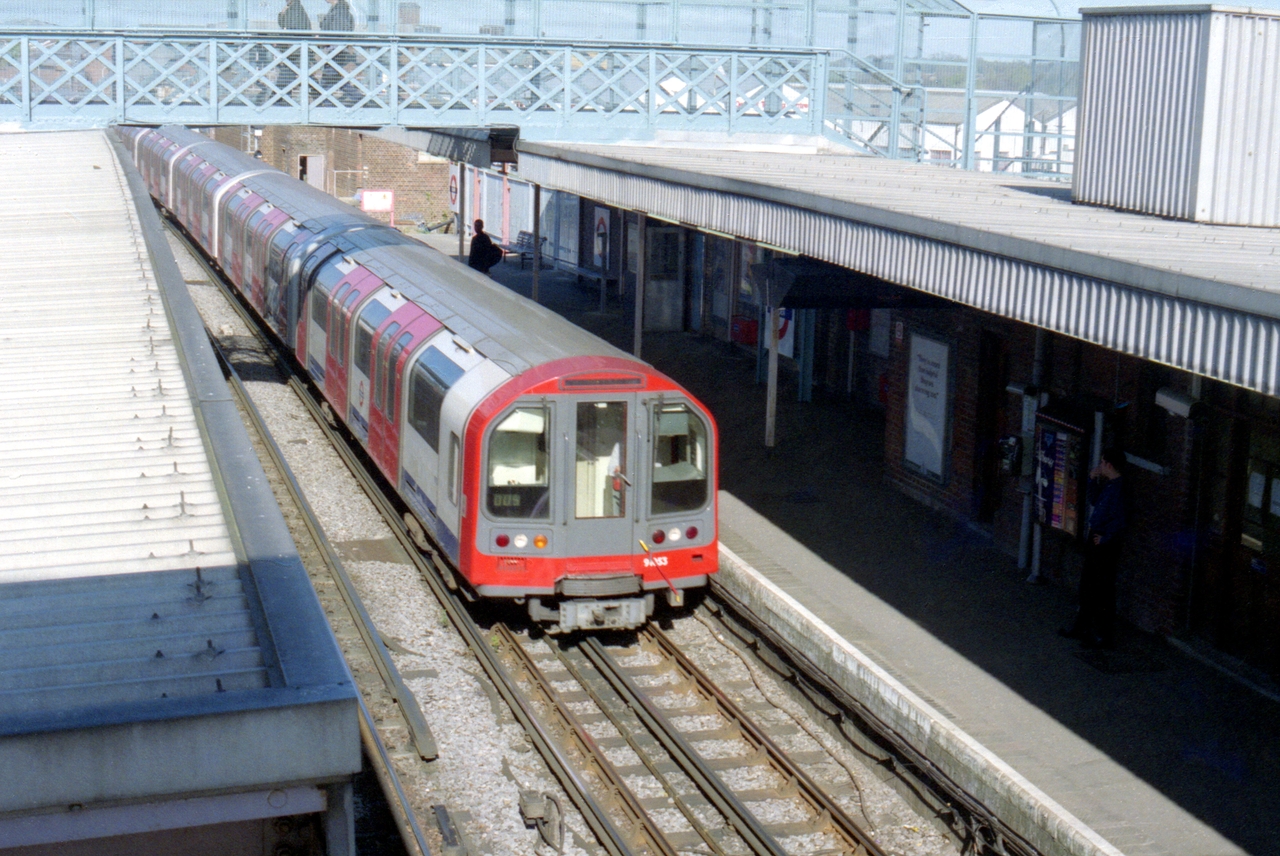}
  \vfill
\end{minipage}\hfill
\begin{minipage}[t]{0.70\linewidth}
\scriptsize
\setlength{\tabcolsep}{2pt}
\renewcommand{\arraystretch}{1.15}

\begin{tabular}{@{}p{0.10\linewidth} p{0.75\linewidth}@{}}

\textbf{BL} &
\begin{tabular}{@{}l p{0.90\linewidth}@{}}
\textbf{A:}  & \textcolor{orange!80!black}{The image does not provide enough information to determine the direction the subway is going relative to the person facing the image.} \textit{(detects ambiguity but does not resolve it)}\\
\end{tabular}
\\[-2pt]
& \rule{\linewidth}{0.2pt} \\[-2pt]

\textbf{CoA} &
\begin{tabular}{@{}l p{0.90\linewidth}@{}}
\textbf{CQ:} & \textcolor{green!50!black}{Is the subway train moving towards or away from you?} \textit{(good: resolves orientation ambiguity)} \\
\textbf{CR:} & Towards me. \\
\textbf{A:}  & \textcolor{green!50!black}{No.} \textit{(good: reasonable answer under given context)} \\
\end{tabular}
\\[-4pt]

\end{tabular}

\end{minipage}
}%
}%
\end{minipage}
\hfill
\begin{minipage}[t]{0.48\textwidth}
\vspace{0pt}
\scriptsize
\colorbox{gray!8}{%
\parbox[t][5.2cm][t]{\linewidth}{%
\vspace{3pt}
\textbf{(f) Q:} The clock is on 12-hour format. What time is it on a 24-hour schedule? \\
\textbf{GTCQ:} Is this photo from the morning or late night?

\vspace{3pt}
\noindent
\begin{minipage}{0.28\linewidth}
  \centering
  \vfill
  \includegraphics[width=\linewidth]{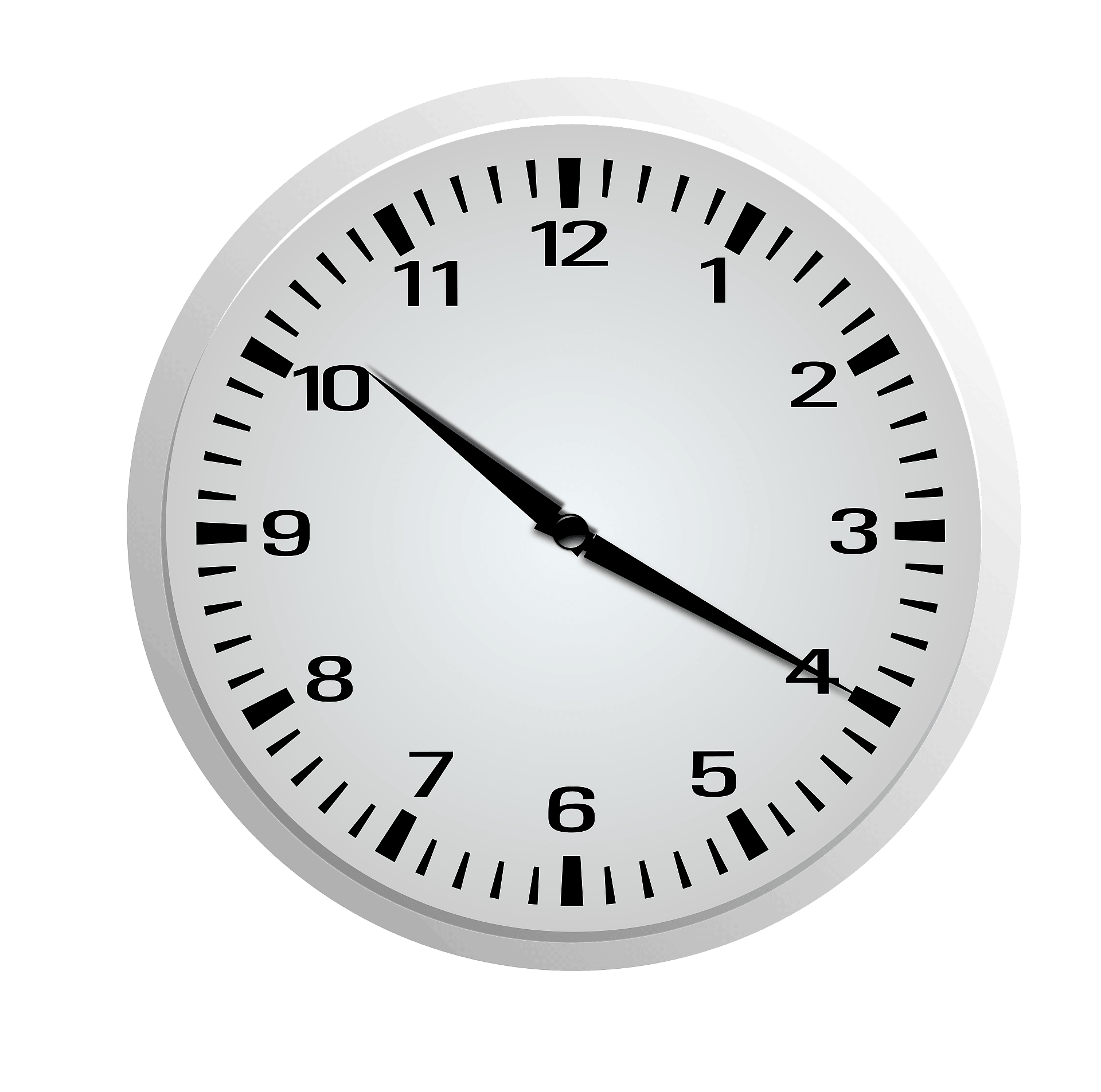}
  \vfill
\end{minipage}\hfill
\begin{minipage}[t]{0.70\linewidth}
\scriptsize
\setlength{\tabcolsep}{2pt}
\renewcommand{\arraystretch}{1.15}

\begin{tabular}{@{}p{0.10\linewidth} p{0.75\linewidth}@{}}
\textbf{BL} &
\begin{tabular}{@{}l p{0.90\linewidth}@{}}
\textbf{CQ:} & \textcolor{orange!80!black}{What is the current time on a 24-hour clock?} \textit{(repeats original question, no additional information gained)} \\
\textbf{CR:} & It is 10:20 on a 24 hour schedule \\
\textbf{A:}  & \textcolor{orange!80!black}{10:20.}  \\
\end{tabular}
\\[-2pt]
& \rule{\linewidth}{0.2pt} \\[-2pt]

\textbf{CoA} &
\begin{tabular}{@{}l p{0.90\linewidth}@{}}
\textbf{CQ:} & \textcolor{green!50!black}{Is the clock showing the time in the morning or afternoon?} \textit{(good:resolves AM/PM)} \\
\textbf{CR:} & In the morning. \\
\textbf{A:}  & \textcolor{green!50!black}{10:20.} \textit{(good: reasonable answer under given context)} \\
\end{tabular}
\\[-4pt]

\end{tabular}

\end{minipage}
}%
}%
\end{minipage}
\hfill

\caption{Case studies illustrating successful and unsuccessful ambiguity resolution. Each block compares the prompting baseline and CoA responses with annotations. Abbreviations: Q = original question; GTCQ = ground truth clarification question; CQ = clarification question; CR = clarification response; A = final answer.}
\label{fig:case_studies_ambiguity}
\end{figure*}

\begin{figure*}[t!]
\centering
\renewcommand{\arraystretch}{1.10}
\setlength{\tabcolsep}{3pt}
\begin{minipage}[t]{0.48\textwidth}
\vspace{0pt}
\scriptsize
\colorbox{gray!8}{%
\parbox[t][4.0cm][t]{\linewidth}{%
\vspace{3pt}
\textbf{(a) Q:} My shop is going to be open. I am the owner of a shop. Should I flip the open sign or keep the position of the sign unchanged? \\
\textbf{GT:} Keep the position of the sign unchanged.

\vspace{3pt}
\noindent
\begin{minipage}{0.28\linewidth}
  \centering
  \includegraphics[width=\linewidth,height=1.2\textwidth,keepaspectratio]{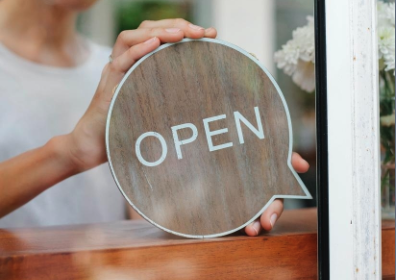}
\end{minipage}\hfill
\begin{minipage}[t]{0.70\linewidth}
\scriptsize
\setlength{\tabcolsep}{2pt}
\renewcommand{\arraystretch}{1.15}

\begin{tabular}{@{}p{0.10\linewidth} p{0.75\linewidth}@{}}

\textbf{BL} &
\begin{tabular}{@{}l p{0.90\linewidth}@{}}
\textbf{A:}  & \textcolor{red}{Flip the open sign.} \textit{(wrong inference)} \\
\end{tabular}
\\[-4pt]
& \rule{\linewidth}{0.2pt} \\[-2pt]

\textbf{CoA} &
\begin{tabular}{@{}l p{0.90\linewidth}@{}}
\textbf{A:}  & Based on the image, it appears that the open sign is already flipped to indicate that the shop is open. \textcolor{green!50!black}{Therefore, you should keep the position of the sign unchanged.} \textit{(reasonable inference)} \\
\end{tabular}
\\[-4pt]

\end{tabular}
\end{minipage}
}%
}%
\end{minipage}
\hfill
\begin{minipage}[t]{0.48\textwidth}
\vspace{0pt}
\scriptsize
\colorbox{gray!8}{%
\parbox[t][4.0cm][t]{\linewidth}{%
\vspace{3pt}
\textbf{(b) Q:} I am driving in New York. Can I make a right turn on the red traffic light? \\
\textbf{GT:} No.

\vspace{6pt}
\noindent
\begin{minipage}{0.28\linewidth}
  \centering
  \includegraphics[width=\linewidth,height=1.2\textwidth,keepaspectratio]{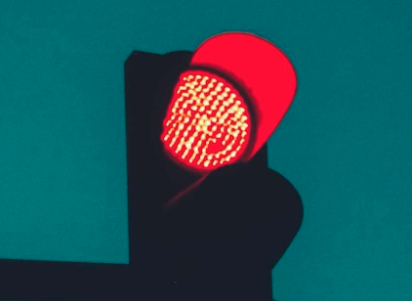}
\end{minipage}\hfill
\begin{minipage}[t]{0.70\linewidth}
\scriptsize
\setlength{\tabcolsep}{2pt}
\renewcommand{\arraystretch}{1.15}

\begin{tabular}{@{}p{0.10\linewidth} p{0.75\linewidth}@{}}
\textbf{BL} &
\begin{tabular}{@{}l p{0.90\linewidth}@{}}
\textbf{A:}  & \textcolor{green!50!black}{No, you cannot make a right turn on a red traffic light in new york.} \textit{(reasonable inference)}\\
\end{tabular}
\\[-2pt]
& \rule{\linewidth}{0.2pt} \\[-2pt]

\textbf{CoA} &
\begin{tabular}{@{}l p{0.90\linewidth}@{}}
\textbf{A:}  & \textcolor{green!50!black}{No, you can not.} The red light typically means "stop" and "do not proceed" in new york. \textit{(reasonable inference)} \\
\end{tabular}
\\[-2pt]

\end{tabular}

\end{minipage}
}%
}%
\end{minipage}

\vspace{4pt}

\caption{Case studies illustrating responses for un-ambiguous questions. Each block compares the prompting baseline and CoA with annotations. Abbreviations: Q = original question; GTCQ = ground truth clarification question; CQ = clarification question; CR = clarification response; A = final answer. }
\label{fig:case_studies_clear}
\end{figure*}